\PassOptionsToPackage{table}{xcolor}
\documentclass[10pt,twocolumn,letterpaper]{article}

\usepackage{cvpr}              
\usepackage{times}
\usepackage{graphicx}
\usepackage{epsfig}
\usepackage{graphicx}
\usepackage{amsfonts}
\usepackage{amsmath}
\usepackage{amssymb}
\usepackage{multirow}
\usepackage{booktabs}
\usepackage[accsupp]{axessibility}
\usepackage{soul}

\usepackage[normalem]{ulem}

\usepackage[dvipsnames]{xcolor}

\def\eq#1{Eq.~(\ref{eqn:#1})}

\newcommand{\mypara}[1]{\vspace{-4mm}\paragraph*{#1}}

\definecolor{cvprblue}{rgb}{0.21,0.49,0.74}
\usepackage[pagebackref,breaklinks,colorlinks,citecolor=cvprblue]{hyperref}

\def\paperID{14733} 
\def\confName{CVPR}
\def\confYear{2024}

\title{Self-Supervised Dual Contouring}

\author{Ramana Sundararaman\\
LIX, Ecole Polytechnique\\
{\tt\small sundararman@lix.polytechnique.fr}
\and
Roman Klokov\\
LIX, Ecole Polytechnique\\
{\tt\small klokov@lix.polytechnique.fr}
\and
Maks Ovsjanikov\\
LIX, Ecole Polytechnique\\
{\tt\small maks@lix.polytechnique.fr}
}
\usepackage[table]{xcolor}

\begin{document}
\maketitle
\begin{abstract}
Learning-based isosurface extraction methods have recently emerged as a robust and efficient alternative to axiomatic techniques. However, the vast majority of such approaches rely on supervised training with axiomatically computed ground truths, thus potentially inheriting biases and data artefacts of the corresponding axiomatic methods. Steering away from such dependencies, we propose a \textit{self-supervised training} scheme to the Neural Dual Contouring meshing framework, resulting in our method: Self-Supervised Dual Contouring (SDC). Instead of optimizing predicted mesh vertices with supervised training, we use two novel self-supervised loss functions that encourage the consistency between distances to the generated mesh up to the first order. Meshes reconstructed by SDC surpass existing data-driven methods in capturing intricate details while being more robust to possible irregularities in the input. Furthermore, we use the same self-supervised training objective linking inferred mesh and input SDF, to regularize the training process of Deep Implicit Networks (DINs). We demonstrate that the resulting DINs produce higher-quality implicit functions, ultimately leading to more accurate and detail-preserving surfaces compared to prior baselines for different input modalities. Finally, we demonstrate that our self-supervised losses improve meshing performance in the single-view reconstruction task by enabling joint training of predicted SDF and resulting output mesh. We open-source our code at \href{https://github.com/Sentient07/SDC}{https://github.com/Sentient07/SDC}.
\end{abstract}    
\section{Introduction}
\label{sec:intro}


Surface mesh extraction from implicit functions, often referred to as isosurfacing~\cite{Lewiner2003,ju2002dual,lorensen1987marching,ManifoldDC}, is a fundamental problem in computer graphics and geometry processing as the quality of reconstructed mesh impacts algorithms used in numerous downstream tasks~\cite{meyer2003discrete}. Prominent primal axiomatic methods such as Marching Cubes~\cite{lorensen1987marching,Lewiner2003}, are computationally efficient but fail to reconstruct sharp features. Dual methods~\cite{ju2002dual,ManifoldDC,SurfaceNets,Kobbelt2001FeatureSS} achieve sharper reconstructions using Quadratic Error Function (QEF) but their performance heavily relies on the data fidelity.

\begin{figure}[t!]
\begin{center}
\includegraphics[width=\linewidth]{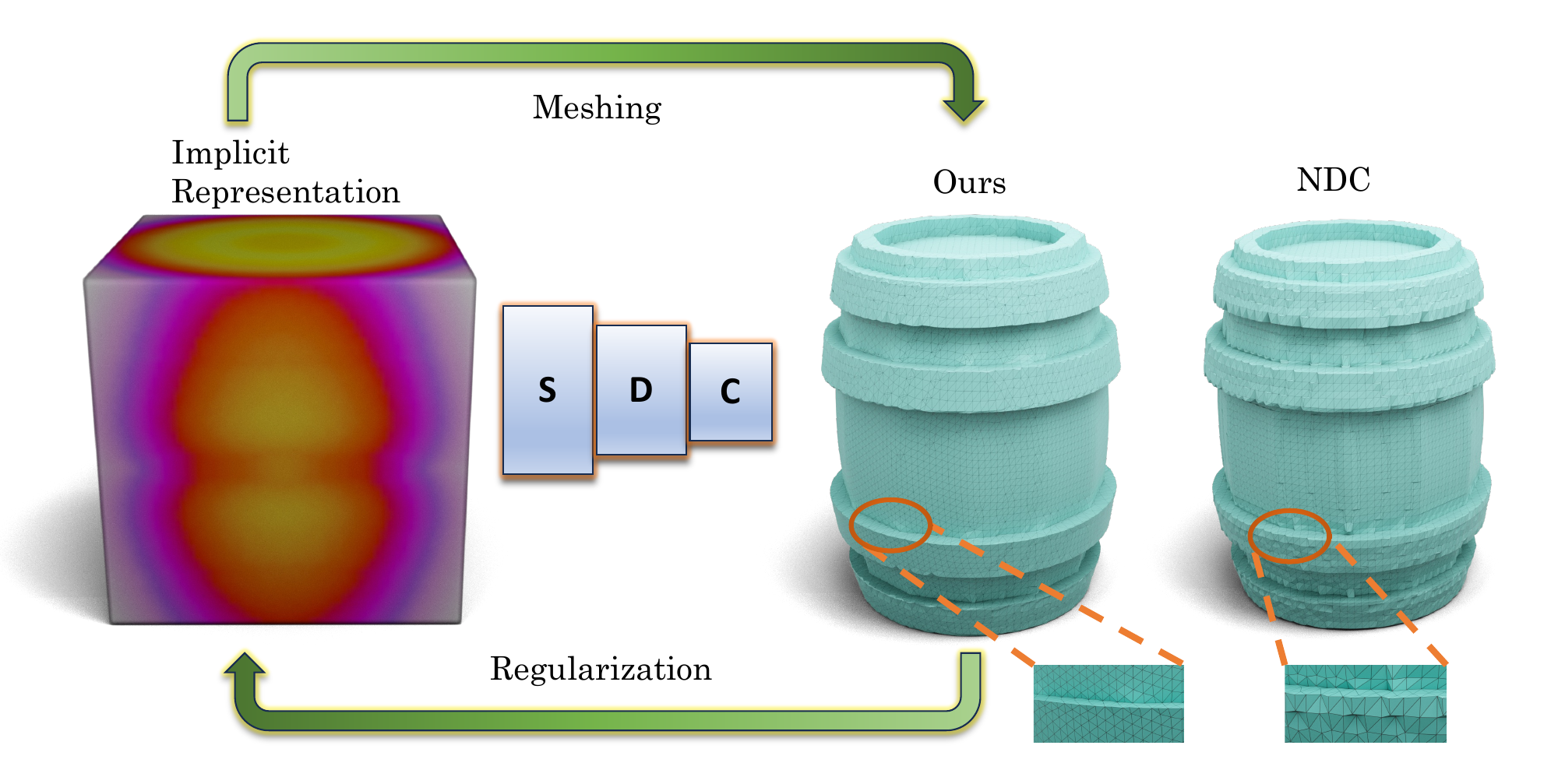}
\end{center}
\vspace{-4mm}
\caption{We propose SDC, a data-driven self-supervised isosurfacing method that reconstructs better feature-preserving meshes from learnt implicit functions compared to the baseline NDC~\cite{chen2022neural}. Our framework is extendable as a regularizer for training Deep Implicit Networks (DINs) for improved reconstruction.
\vspace{-3mm}}
\label{fig:teaser}
\end{figure}


Following the ubiquitous success of neural methods in 2D and 3D applications, recent learning-based meshing approaches~\cite{chen2021neural,chen2022neural} propose to predict mesh vertices and connectivity from various inputs in a feed-forward manner. When trained on extensive, augmented mesh collections, these methods demonstrate enhanced generalization and yield higher-quality reconstructions. However, these approaches rely either directly on the ground truth meshes~\cite{sharp2020pointtrinet} or on some intermediate pre-computed proxy meshes, like complex tessellation templates~\cite{chen2021neural} and outputs of axiomatic approaches~\cite{chen2022neural}. 

Neural Dual Contouring (NDC)~\cite{chen2022neural}, the most relevant method to our work, is a promising recent approach that enhances standard Dual Contouring (DC)~\cite{ju2002dual,ManifoldDC} with a data-driven pipeline. It aims to replicate DC-produced meshes, resulting in a framework capable of generating high-quality, feature-preserving meshes. However, since vertex estimation of DC follows QEF minimization, it requires precise SDF values at grid points and normals at edge intersections for training. Due to QEF minimization being ill-defined in particular scenarios~\cite{ju2002dual}, and susceptible to irregularities in the input SDF, NDC may exhibit a bias towards poor triangulation quality for challenging inputs and intricate details. Furthermore, its reliance on DC meshes for training restricts its use in an end-to-end scenario where both the implicit function and the explicit mesh can be optimized jointly.

To overcome these limitations, we propose Self-Supervised Dual Contouring (SDC), a meshing framework that extends Neural Dual Contouring with two novel geometrically motivated self-supervised losses to learn the mesh vertex positions from SDF inputs while relying on sign changes to extract connectivity. The first self-supervised objective is a distance-based loss which minimizes the difference between the input implicit function values at points close to the surface and distances from the same points to the reconstructed mesh. This ensures that the predicted reconstructed mesh agrees with the input SDF. Secondly, we align the normals corresponding to reconstructed faces to those \emph{estimated} from the input grid of SDFs. Since our training objective relies \emph{only} on the input SDF and is agnostic to any explicit representation of the shape, e.g., based either on surface samples~\cite{Point2Mesh,sharp2020pointtrinet} or on mesh vertices~\cite{chen2021neural,chen2022neural}, we refer to our approach as \emph{Self-Supervised}. Together these two terms produce vertices that best explain the input grid of signed distance function values up to the first order like dual methods, while avoiding shortcomings of the quadratic error function (QEF) minimization~\cite{ju2002dual,chen2022neural}. 
In particular, unlike DC, our method produces vertices that lie within each voxel-cell, and optimization of our self-supervised objectives leads to overall better results for input data with imperfections. As a result, we demonstrate better performance than supervised baselines in obtaining sharper meshes with negligible self-intersections and better generalization to unseen input data.

In addition, leveraging the connection between reconstructed meshes and input SDF grids, we propose a mesh-based \textit{regularization} for training Deep Implicit Networks (DINs) - MLPs which represent zero-level sets of a shape. In a standard setup, training of DINs is completely decoupled from meshing, which is only performed a posteriori. Thus, existing DINs might produce implicit functions that do not correspond to a distance function arising from an extracted mesh. This motivates us to use SDC as a regularization to train DINs. In particular, we enforce the distance function produced by DINs during training to remain as close as possible to the distance function arising from the mesh produced by SDC. This forms the \textit{converse} of our self-supervised meshing losses, but now geared to produce more coherent \textit{implicit functions}. We demonstrate that this regularization can be directly adopted into existing training paradigms for Deep Implicit Networks and it that leads to more plausible reconstructions.

Finally, since SDC training does not require ground truth meshes, it can be seamlessly integrated within end-to-end training paradigms. To highlight this, we consider the task of mesh reconstruction from images where we first construct an implicit representation given an image using an existing approach~\cite{park2019deepsdf,remelli2020meshsdf} and then reconstruct a mesh using SDC. The resulting model is trained end-to-end by \textit{jointly} minimizing the standard surface reconstruction loss, our self-supervised loss and the proposed regularization for Deep Implicit Networks. The resulting approach demonstrates compelling surface reconstruction efficacy across multiple object categories from the ShapeNet dataset~\cite{shapenet2015}. In summary, our contributions are as follows:
\begin{itemize}
\item  We extend Neural Dual Contouring with two novel self-supervised loss functions ensuring consistency between input SDF and distance to output meshes, which is geometrically better motivated than the standard QEF mininmization.
\item We introduce a novel regularizer for training Deep Implicit Networks (DINs), by penalizing SDFs that do not correspond to underlying meshes.
\item We show that SDC generalizes better than supervised meshing methods, and further demonstrate its utility in feature-preserving mesh reconstruction from images.
\end{itemize}

\section{Related work}
\label{sec:related}

Surface reconstruction and meshing tasks have been extensively studied within computer graphics and vision. Below, we review the methods that are most closely related to ours and refer interested readers to several surveys~\cite{khatamian2016survey,shewchuk2016delaunay,berger2017survey} for a more in-depth discussion.

\mypara{Axiomatic contouring techniques.} The process of iso-surface extraction from volumetric data is referred to as contouring. The most widely-used pipeline for surface reconstruction consists of two major steps. First, volumetric data is computed using a signed distance function ~\cite{hoppe1992surface,curless1996volumetric,kazhdan2006poisson,mullen2010signing}. In the second step, a mesh is extracted from this representation, using Marching Cubes \cite{lorensen1987marching}, Dual Contouring \cite{ju2002dual} or related approaches \cite{doi1991efficient,treece1999regularised,newman2006survey}. This pipeline can produce very high quality meshes, but is sensitive to the quality of input data and is not differentiable and thus does not easily fit within modern learning-based pipelines~\cite{liao2018deep}.
Another line of approaches is based on Delaunay triangulations,
\cite{boissonnat1984geometric,kolluri2004spectral,boissonnat2005provably}, as well as closely-related constructions~\cite{edelsbrunner1994three,bernardini1999ball,amenta1998new,amenta2001power} which often come with strong theoretical guarantees and are typically geared towards preserving some input point set~\cite{bernardini1999ball,amenta1999surface}. Such methods, however, provide little control over the output triangulation. Moreover, classical approaches are typically not differentiable and thus cannot be easily integrated into learning pipelines.

\mypara{Data-driven mesh reconstruction.} To address differentiability issues of classical mesh reconstruction approaches, a number of data-driven techniques have recently been proposed. This includes differentiable variants of contouring methods, such as the Marching Cubes \cite{liao2018deep,chen2021neural}, dual Marching Cubes~\cite{FCubes}, Marching Tetrahedra \cite{shen2021deep}, and Dual Contouring \cite{chen2022neural}. Our approach can be categorized into this family of differentiable contouring techniques. In addition, surface reconstruction has also been tackled using template-based techniques which fit a template mesh to the input \cite{litany2018deformable,kanazawa2018learning,lin2019photometric}, or deform an initial mesh while potentially updating its connectivity \cite{wang2018pixel2mesh,pan2019deep,gao2020learning}. 
Furthermore, other local approaches propose to fit parameterized surface patches to points \cite{groueix2018papier,williams2019deep}, or to decompose space into convex sets \cite{deng2020cvxnet,chen2020bsp,maruani23iccv}. Finally, several learning-based methods mesh an input point set~\cite{dai2019scan2mesh,sharp2020pointtrinet,liu2020meshing,rakotosaona2021learning}. These approaches strongly rely on the given input point cloud and can thus be sensitive to artifacts and noise.

\mypara{Neural fields.} There has been a recent surge in methods for learning implicit functions, referred to as neural fields~\cite{xie2022neural}. This includes pioneering methods for predicting grid occupancy and signed distance functions \cite{mescheder2019occupancy,chen2019learning,park2019deepsdf} as well as their many follow-up works \cite{genova2019learning,saito2019pifu,sitzmann2020implicit,tancik2020fourier}. Often, training Deep Implicit Networks introduces inductive bias~\cite{pumarolavisco} thereby producing inexplicable surface behaviors. Multiple regularization techniques~\cite{IGR,Lipman2021PhaseTD,pumarolavisco,LipRegul,sitzmann2020implicit,tancik2020fourier,lindell2021bacon,ben2022digs} have been proposed to address that issue. They largely focus on either learning high-frequency signals~\cite{tancik2020fourier,lindell2021bacon} or designing loss functions based to guide the gradient level-set~\cite{IGR,Lipman2021PhaseTD,pumarolavisco}. While the latter category of methods has convergence guarantees~\cite{Lipman2021PhaseTD}, the former helps in producing detail preserving reconstruction. Our work, on the other hand, provides a \emph{novel} regularization by utilizing a meshing framework to ensure that the predicted signed distance field arises from a mesh.
Similar to recent works~\cite{mehta2022level,remelli2020meshsdf,guillard2022udf,peng2021shape,shen2021deep,yang2021geometry,FCubes} which optimize the mesh by controlling the underlying implicit function, our approach also enables back-propagation from the mesh to the input SDF. However, our approach is non-iterative and produces a mesh in one feed-forward pass.

\section{Background}
We first begin by providing the notations used across the paper in Section~\ref{sec:notation} and review the Dual Contouring and Neural Dual Contouring algorithms in Section~\ref{sec:DC}. 

\subsection{Notations}
\label{sec:notation}
We let $f: \mathbb{R}^3 \rightarrow \mathbb{R}$ be an implicit function that determines the signed distance of a query point $x \in \mathbb{R}^3$ from an underlying surface $\chi$. We let $f_{\theta}$ denote the implicit function produced by a neural network parameterized by $\theta$. We use $\mathcal{G}$ to denote an $l\times m \times n$ regular grid in the Euclidean space, $g \in \mathcal{G}$ to be a node in the grid and $e_{ij} \in \mathcal{G}$ to be an edge between adjacent nodes $g_i,g_j$ respectively. We denote $\mathcal{S}_\mathcal{G}$ as the discretization of $f$ on $\mathcal{G}$ \ie SDF evaluated at every grid node. Finally, we let $\lambda$ denote the latent vector associated with a shape computed using an encoder.  

\subsection{Dual contouring}
\label{sec:DC}
The Dual Contouring (DC)~\cite{ju2002dual} framework is designed to produce quadrilateral meshes from input signed distance functions discretized on a grid $\mathcal{S}_\mathcal{G}$. It can be decomposed into two main steps: (1) construction of mesh faces $\mathcal{F}$; (2) computation of mesh vertices $\mathcal{V}$. To construct mesh faces, DC considers adjacent nodes $g_i, g_j \in \mathcal{G}$ such that their signed distance are of opposite signs, $f(g_i) \neq f(g_j)$. Then, for those pairs, a unique quadrilateral face $q_k$ is constructed such that $q_k$ crosses the edge $e_{ij}$. Vertices of this quadrilateral $q_k$ reside in adjacent grid cells. The faces of the resulting mesh are obtained as a union of all quadrilateral faces $\mathcal{F} = \bigcup_{k \in I} q_{k}$, where $I$ is a set of indices of grid edges connecting grid vertices of different signs. Final mesh vertex positions within each cell are obtained by minimizing the Quadratic Error Function (QEF). More precisely, given $p_e$ to be the intersection position between edge $e_{ij}$ and the face $q_{ij}$ (along $e_{ij}$) and $n_e$ to be the gradients of the SDF at positions $e_{ij}$ (or surface normals), vertices are determined for each grid cell $c$ by solving:
\begin{equation}
v_c = \underset{x}\arg\min \sum_{e \in I_c} [n_e * (x - p_e)],
\label{eqn:QEF}
\end{equation}
where $I_c$ is a set of indices of intersected grid edges corresponding to cell $c$. Recently introduced Neural Dual Contouring~\cite{chen2022neural} attempts to emulate Dual Contouring via a neural network in order to reduce the inference time while achieving optimal reconstruction.

\section{Method}
\label{sec:method}
We describe our Self-supervised Dual Contouring (SDC) loss functions in Section~\ref{sec:Recon}, discuss how to use them for regularization while training deep implicit networks in Section~\ref{sec:Regul}, and apply them for mesh reconstruction from input images in Section~\ref{sec:SVR}.

\subsection{Self-supervised dual contouring (SDC)}
\label{sec:Recon}
We introduce a Self-supervised training scheme for Dual Contouring, and refer to a model trained with it as to SDC. It excludes the dependence on vertex positions obtained via QEF minimization. SDC produces quad meshes from the signed distance function discretized on a grid $\mathcal{S}_\mathcal{G}$ similarly to any Dual Contouring method. We triangulate quadrilaterals by joining the top-left and bottom-right vertex as a convention. $\mathcal{S}_\mathcal{G}$ can either be measured~\cite{Hart1996} or predicted using a DIN~\cite{park2019deepsdf} $f_{\theta}$. SDC consists of two main components constructing faces and vertices. To construct faces, we follow the Dual Contouring~\cite{ju2002dual} algorithm to determine the connectivity based on signed distance value at grid nodes. A quadrilateral face is constructed in the cells incident on every edge $e_{ij} \in \mathcal{G}$ connecting grid nodes of different signs $\mathrm{sign}(f(g_i)) \neq \mathrm{sign}(f(g_j))$. For vertex prediction, we use a 6-layered 3D Convolutional Neural Network $h_\phi: \mathcal{S}_\mathcal{G} \rightarrow \mathcal{P}\in \mathbb{R}^{(l-1)\times(m-1)\times(n-1)\times3}$, which takes the grid of signed distance values as input and predicts a single point per each grid cell as the output. Then, we apply a masking to select grid cells, that bear node SDF values of opposite signs. Please refer to Suppl. for more details.

SDC is trained without explicit fitting to the ground truth meshes using two loss functions. Firstly, to ensure formal correspondence of produced surfaces to input SDF grids, we propose a distance-based loss $\mathcal{L}_{D}$. This objective minimizes the differences between 1) the absolute distance values at the nodes of the discretization grid $f(g_i) \in \mathcal{S}_{\mathcal{G}}$ which are provided as inputs and 2) the distance $\mathbf{d}^{p}(g_i, \mathcal{M})$ \emph{measured} from the same point $g_i$ to the \textit{predicted} mesh $\mathcal{M}$. For a smooth or gently undulating surface, this distance provides a reasonable approximation of the underlying geometry~\cite{LocRegSGP,Pottmann2003}. However, in the presence of sharp features, the local geometry changes direction rapidly. Close query points residing on different sides of a sharp feature have similar distances to the mesh but may be projected to very different locations on the actual surface. To address this, we also introduce a normal consistency loss, which aims to align the normals estimated at the nodes of the discretization grid $f(g_i) \in \mathcal{S}_{\mathcal{G}}$ and normals corresponding to the generated mesh face. 

\begin{figure}[ht]
  \centering
  \begin{subfigure}{.5\columnwidth}
    \includegraphics[width=\linewidth]{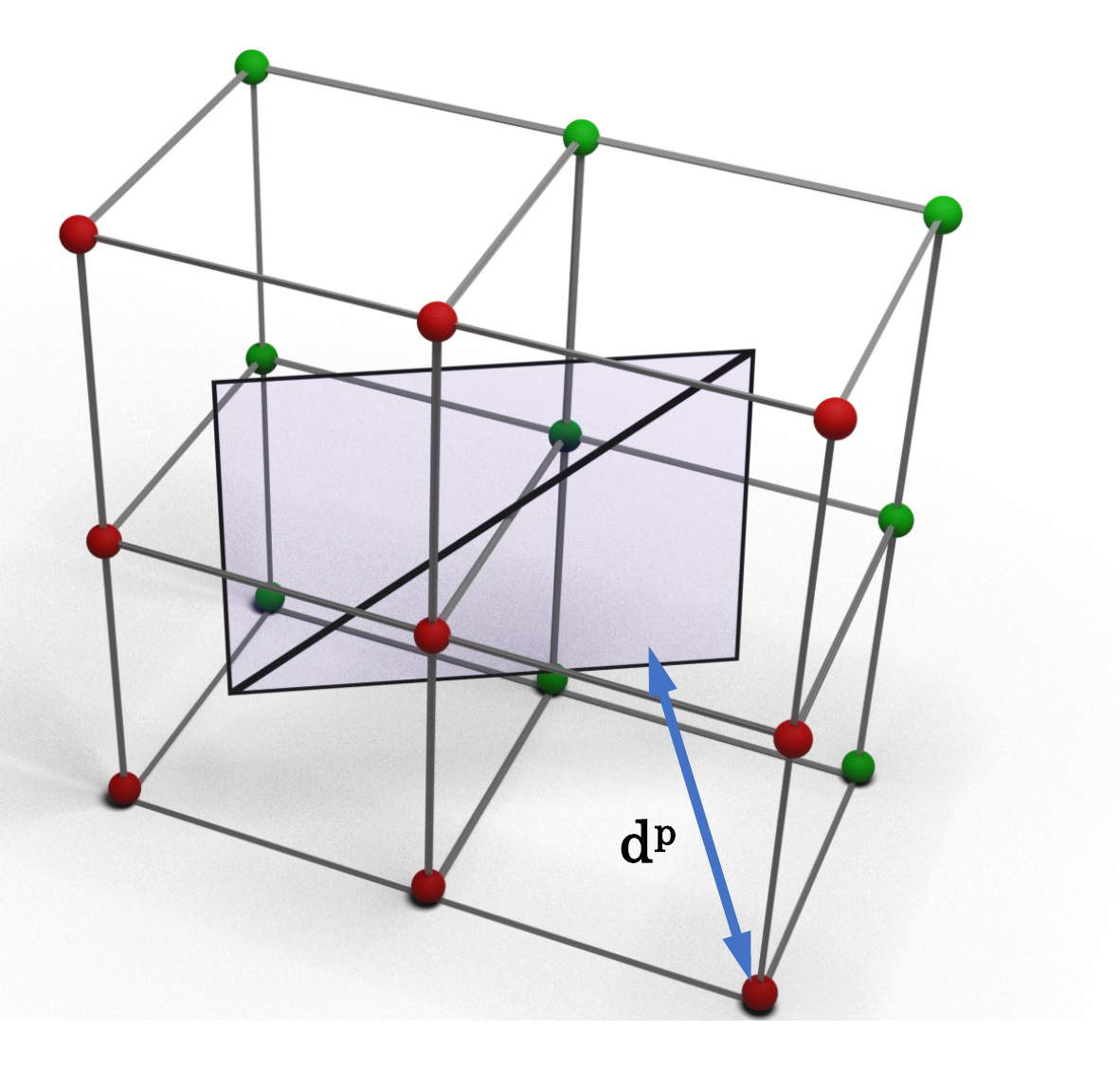}
    \caption{}
  \end{subfigure}
  \hfill
  \begin{subfigure}{.49\columnwidth}
    \includegraphics[width=\linewidth]{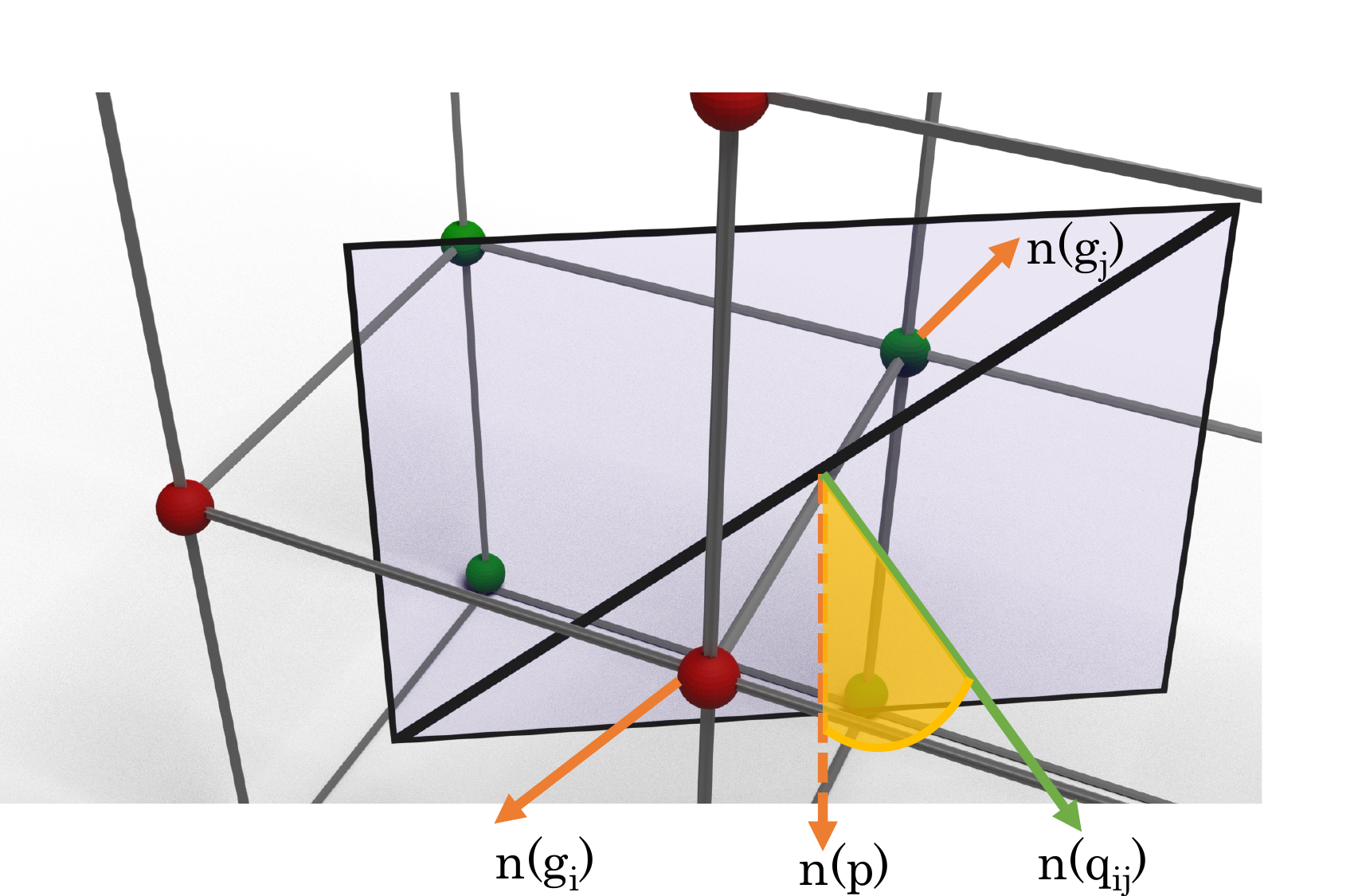}
    \caption{}
  \end{subfigure}
  \caption{Illustration of our self-supervised losses. (a) Measurements of $\mathbf{d}^{p}$ between grid nodes and the closest point on a mesh face. For visualization purposes, we only show this estimation at one node. (b) Normal consistency loss is measured as the discrepancy between interpolated normals (orange) and the face normal (green). Red nodes denote negative SDF grid and green denotes positive. The constructed quad is shown in transparent purple.}
  \label{fig:method}
\end{figure}

We refer to the losses proposed above as \textit{self-supervised} since the training signal for our pipeline comes from the input alone. We do not use any explicit surface discretization for training and our losses are solely based on the input implicit function values and the generated mesh. In contrast, we refer to methods that predict vertices to fit a specific mesh structure as supervised meshing methods~\cite{chen2021neural,chen2022neural}. We provide visual explanations of our loss functions in Figure~\ref{fig:method}. In summary, our combined self-supervised objective for the vertex prediction network is given by:
\begin{equation}
\mathcal{L}_{\mathrm{Mesh}} = \mathcal{L}_{D} + \alpha_1 \mathcal{L}_{N}, 
\label{eqn:LUnsup}
\end{equation}
where $\alpha_1$ is a hyperparameter to weigh $\mathcal{L}_{N}$. 

\paragraph{Distance loss.} 

The objective of our distance loss is to minimize the discrepancy between surfaces produced by SDC and zero-level sets defined by the input SDFs. In order to achieve this, given grid nodes $g_i \in \mathcal{G}$, we measure the distance between the grid nodes and the mesh produced by SDC and penalize its deviation from the input absolute distance values $f(g_i)$. In summary, our distance loss is defined as:
\begin{equation}
    \mathcal{L}_{D} = \sum_{g_i \in \mathcal{G}}||\mathrm{abs}(f(g_i)) -  \mathbf{d}^{p}(g_i, \mathcal{M})||^2_{2},
\label{eqn:sdfr}
\end{equation}
where $\mathbf{d}^{p}$ is the point-to-face distance, computed between $g_i$ and its projection on $\mathcal{M}$. For computational efficiency, we compute $\mathcal{L}_{D}$ only at grid nodes that are close to the surface. Please refer to Suppl for details.

\paragraph{Normal consistency loss.} 
We introduce a novel normal consistency loss, $\mathcal{L}_N$ which aims to align the level-sets up to the \emph{first order}. In particular, given the grid of scalar SDFs, we compute normals as gradients $n(g_i) := f'(g_i) / ||f'(g_i)||_2$ in grid nodes $g_i$ using Five-point stencil, a commonly used finite difference approach for solving partial differential equations on regular grids~\cite{SmithPDE}. More details on gradient estimation is provided in the Suppl. Given an edge $e_{ij} \in \mathcal{G}$ connecting nodes $g_i, g_j$ of the opposite signs, we first estimate a point $p \in e_{ij}$ along the edge where $f(p)=0$. The point $p$ can be represented in terms of $g_i$ and $g_j$  as $p = (1-t)g_i + t g_j$. where $t$ is a parameter that determines the position of $p$ along the edge $e_{ij}$. Solving for $p = 0$, the parameter $t$ can be expressed in closed form as:
\begin{equation}
t = \frac{{f(g_i)}}{{f(g_i) - f(g_j)}}.
\end{equation}

Similarly, we can estimate the normal at the point $p$ by linear interpolation. Using the previously computed parameter $t$, the normal at $p$ can be represented as:
\begin{equation}
n(p) = \mathrm{sign}(f(g_i))(1-t)n(g_i) + \mathrm{sign}(f(g_j))tn(g_j).
\end{equation}
This expression combines the known gradients at the neighboring nodes, weighted by the parameter $t$, to estimate the gradient at a point $p$. Since the inward normal is oriented opposite to the outward normal, we multiply by the $\mathrm{sign}$ for consistency. As our mesh is constructed following Dual Contouring~\cite{ju2002dual}, we denote the quad-face constructed dual to the edge $e_{ij}$ as $q_{ij}$ and its normal as $n(q_{ij})$. With the aim of aligning these two normals, our normal consistency loss can be written as:
\begin{equation}
\mathcal{L}_{N} = 1 - \frac{{n(q_{ij}) \cdot n(p)}}{{\|n(q_{ij})\| \|n(p)\|}}.
\end{equation}

\paragraph{Training augmentation.}
At training time, we add synthetic noise which is zero-mean Gaussian noise with a standard deviation equivalent to a third grid edge length $e_{ij}$ to the input SDF. Since SDC constructs mesh faces based on Dual Contouring, a noise that induces a sign change to the initial SDF will also lead to a topological change. More precisely, augmentation will lead to inclusion of vertices (and faces) in cells whose nodes differ in sign. Similarly, faces are removed dual to cells whose signs agree after augmentation. More training details are provided in the Suppl.

\subsection{Deep implicit network regularizer}
\label{sec:Regul}
Deep Implicit Networks (DINs), commonly represented by Multi-Layer Perceptrons (MLP), are trained to acquire an implicit representation of a shape and are primarily utilized for various inference-based surface reconstruction tasks. In a traditional setup, DINs are trained by direct supervision of the SDF values at query points sampled close to the surface. Such a training scheme, as also observed by previous works~\cite{IGR,Lipman2021PhaseTD,pumarolavisco}, is known to produce inexplicable behavior, such as ambiguous level sets. To address this, we propose a novel mesh-based training regularization for DINs to ensure that the Distance Field (DF) produced by DIN agrees well with the resulting extracted mesh. Our regularization minimizes the discrepancy between the predicted DF (absolute value of SDF) and the distance computed from the mesh corresponding to the SDF extracted using SDC.

We assume that we are endowed with a learnable DIN $f_\theta (\lambda, p): p\in\mathbb{R}^{3} \rightarrow s_{p} \in \mathbb{R}$, represented as an MLP, whose weights $\theta$ encode the implicit surface of a shape conditioned by a latent vector $\lambda$, $p$ is a point in space such that $f_\theta(\lambda, p) = s_{p}$ is its signed distance from the zero-level set. Then, for a given $\mathcal{S}_{G}$ - a discretization of $f_\theta$ on a regular grid and $\mathcal{M}^{*}(\mathcal{S}_\mathcal{G})$ - a mesh that is produced following any iso-surfacing algorithm, it is not necessarily true that $\mathrm{abs} (f_\theta (p, \lambda)) = \mathbf{d}^p (p, \mathcal{M}^{*}(\mathcal{S}_\mathcal{G}))$, where $\mathbf{d}^p$ is the point-to-face distance. While this disparity is viewed as an inherent drawback of primal iso-surfacing methods~\cite{DMC}, for dual methods, this disparity can, in principle, be minimized due to their ability to preserve sharp details in the reconstructed meshes. This motivates us to use our SDC for iso-surface extraction and regularize the training of DINs by minimizing the aforementioned disparity. Assuming we have access to a dataset consisting of $N$ shapes and a pre-trained SDC with frozen weights, our regularization objective can be summarized as follows:
\begin{equation}
\mathcal{L}_{SDR} = \sum_{j=1}^N \sum_{g_i \in \mathcal{G}} ||\ \mathrm{abs}(f_\theta(\lambda_j, g_i) - \mathbf{d}^{p}(g_i, \mathcal{M})\ ||^2_{2}.
\label{eqn:sdr}
\end{equation}
This regularization penalizes the discrepancy between the DF at grid nodes predicted by the DIN $f_\theta(g_i)$ and the distance $\mathbf{d}^p$ computed between the grid nodes and the mesh $\mathcal{M} = \mathrm{SDC}(\mathcal{S}_\mathcal{G})$. To obtain $\mathcal{M}$, we first evaluate $f_{\theta}$ over a discrete grid $\mathcal{G}$ and obtain $\mathcal{M} = \mathrm{SDC}(\mathcal{S}_\mathcal{G})$. Note that the above regularization is similar to the signed distance loss defined in \eq{sdfr}, but with an important difference. Here we do not use ground-truth SDF values but instead, regularize the \emph{predicted} signed distance. This regularization is applied alongside the main objective function to reconstruct the implicit surface commonly used to train DINs~\cite{park2019deepsdf,CSDF} as follows:
\begin{equation}
  \mathcal{L}_\mathrm{SDF} = \sum_{j=1}^N \sum_i^{K} ||\left(f_\theta\left(\lambda_j, x_i\right) - s_i\right||^2_2+\frac{1}{\sigma^2}\left\|\lambda_j\right\|^2_2,
\label{eqn:DINSDR}
\end{equation}
where K is the number of points with annotated SDF values per shape $j$, $s_i$ is the ground-truth SDF value, and $\sigma$ is a parameter used to promote compactness in latent space~\cite{park2019deepsdf}.

Combining the standard training loss for DINs~\cite{park2019deepsdf} alongside our regularization weighted with a scalar $\alpha_2$, our learning objective is formulated as follows:
\begin{equation}
    \mathcal{L}_\mathrm{train} = \mathcal{L}_\mathrm{SDF} + \alpha_2 \mathcal{L}_{SDR}.
\label{eqn:DINTr}
\end{equation}
At the inference time, we reconstruct surfaces from point clouds by first recovering the optimal latent vectors (See Eq. (4) in~\cite{CSDF}). Then, we predict a grid of SDF values $\mathcal{S}_\mathcal{G}$ and extract the iso-surface using our SDC.

\subsection{Joint learning of implicit surface and meshing}
\label{sec:SVR}
Differently from existing data-driven meshing methods, SDC relies neither on explicit mesh data~\cite{rakotosaona2021learning,sharp2020pointtrinet} nor on axiomatically produced proxy meshes~\cite{chen2022neural,chen2021neural}. This means that SDC can be used to optimize both SDF and reconstructed meshes jointly in an end-to-end manner. To demonstrate this we consider the task of predicting surface meshes from images. We first model the SDF representation of a shape given an image and then predict the mesh using SDC. In terms of the architectures, we follow MeshSDF~\cite{remelli2020meshsdf}: we produce latent vectors $\lambda$ from input images with a ResNet-18~\cite{Resnet} encoder and learn an implicit shape representation of shapes using DeepSDF~\cite{park2019deepsdf} conditioned on the aforementioned latent vectors. Instead of performing test-time optimization like MeshSDF, we predict SDF values on a regular grid $f_\theta(\mathcal{G})$ and use SDC to reconstruct a mesh $\mathcal{M}=\mathrm{SDC}(\mathcal{S}_\mathcal{G})$ by a simple forward pass of our SDC network. Initially, we train the networks (1) DeepSDF (with an image encoder) and (2) SDC independently, then jointly fine-tune them for 200 epochs. This fine-tuning uses an image as input and the mesh $\mathcal{M}$ from SDC as output, while minimizing the combined objective terms for surface reconstruction and meshing as follows:
\begin{equation}
    \mathcal{L}_{\mathrm{SVR}} = \mathcal{L}_\mathrm{SDF} + \alpha_3 \mathcal{L}_{SDR} + \alpha_4 \mathcal{L}_\mathrm{D},
\end{equation}
where $\mathcal{L}_\mathrm{SDF}$ and $\mathcal{L}_\mathrm{SDR}$ (defined in \eq{DINSDR} and \eq{sdr} respectively) are used to reconstruct and regularize the implicit representation, while $\mathcal{L}_{D}$, defined in \eq{sdfr}, is used to train SDC. Since the evaluated SDF might be noisy during training, we observed that using $\mathcal{L}_{N}$ results in instability.
\section{Experiments, results and discussion}
We show the effectiveness of SDC across three tasks, namely, implicit function meshing, DIN training regularization and joint SDF and mesh prediction from images.

\paragraph{Evaluation Metrics.}
We use Chamfer Distance (CD), Normal Consistency (NC), Self-Intersection (SI), Edge Chamfer Distance (ECD), 3D-IoU, Precise Level-Set Discrepancy (LSD-P), and Approximate Level-Set Discrepancy (LSD-A) as metrics to compare all approaches. For measuring Level-Set discrepancy, we use two terms, Precise (LSD-P) and Approximate (LSD-A). The former is used in case where shapes are water-tight and analytical SDF can be measured. In such cases, we measure the discrepancy between the ground truth analytical SDF and the distance function measured from the generated mesh. LSD-A is used for non-watertight shapes and is measured by sampling points on the ground truth mesh and measuring its distance from the generated mesh's faces. Self-intersection is reported as the total number of intersecting triangles per mesh averaged across the test set. For the single-view reconstruction task, we also measure the Structural Similarity Index (SSIM) between the rendering of the reconstructed mesh and the ground truth mesh. We measure 3D-IoU for experiments where the generated mesh could topologically differ from the ground truth. CD, ECD, LSD are scaled by $10^3$ while NC, SSIM 3D IoU are in \%. We provide more details on all the metrics in the Suppl. 

\begin{figure}[h]
  \includegraphics[width=\linewidth]{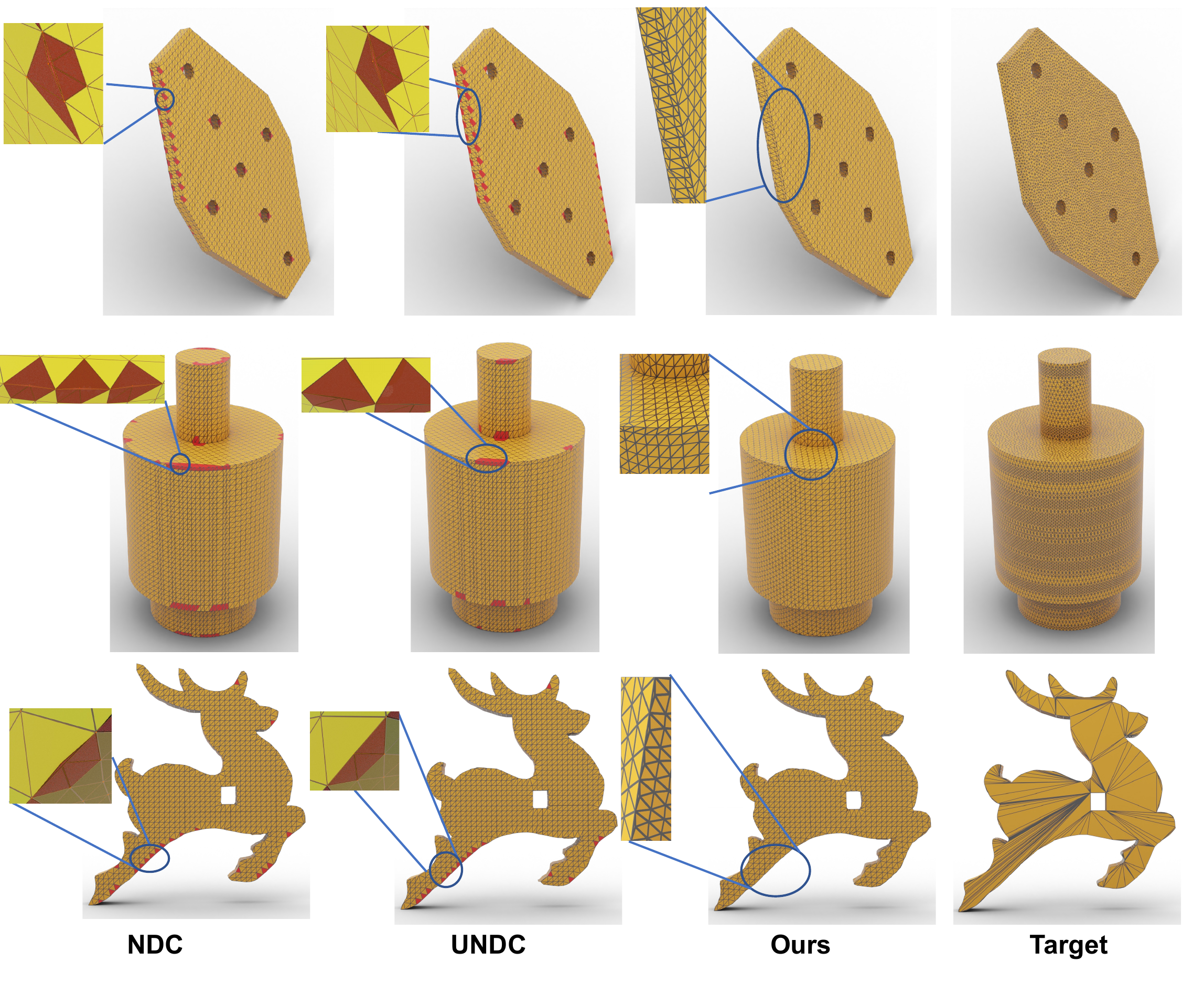}
  \caption{We show three qualitative mesh reconstruction examples from the ABC dataset (Row 1,2) and the Thingi10k dataset (Row 3). Self-intersecting faces are highlighted in red. QEF-based learning methods NDC and UNDC show significantly higher self-intersecting faces along sharp edges while ours does not. }
\label{fig:NDCvsOurs}
\end{figure}

\begin{figure}[ht]
\begin{center}
  \includegraphics[width=\linewidth]{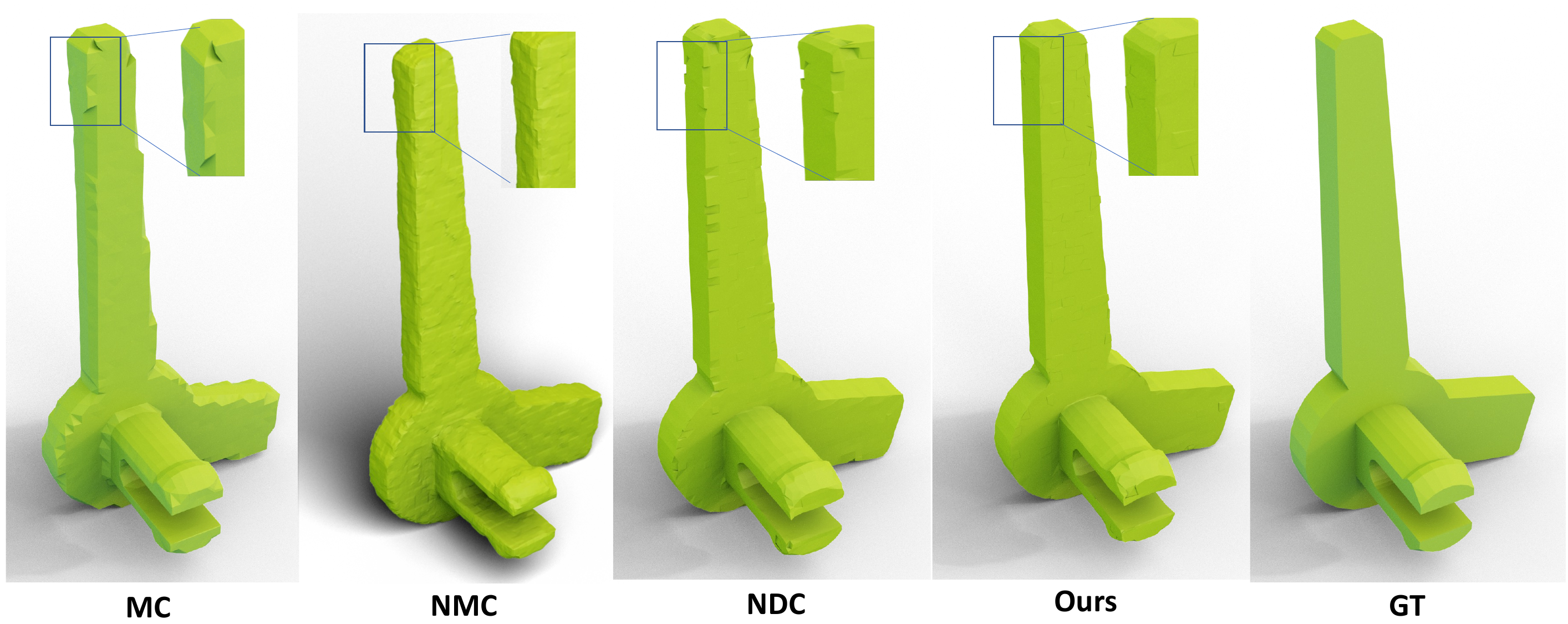}
  \caption{Qualitative examples of meshing SDF produced by NGLOD~\cite{takikawa2021nglod}. MC, NMC, and NDC produce noisy meshes while SDC produces a smoother yet feature-preserving reconstruction.}
\label{fig:LearntSDF}
\end{center}
\end{figure}

\begin{table}[h]
\centering
\resizebox{\columnwidth}{!}{%
\begin{tabular}{ccllllll}
\hline
\rowcolor{gray!25}
Dataset & \multicolumn{1}{l}{Input SDF } & Method                     & CD ($\downarrow$) & NC ($\uparrow$) & $\#$SI ($\downarrow$) & ECD ($\downarrow$) & LSD-P ($\downarrow$) \\ \hline
\multirow{6}{*}{ABC}        & \multirow{6}{*}{Analytical} & MC~\cite{Lewiner2003}      & 4.7               & 92.1            & \textbf{0}                 & 4.8                & 16.1               \\
                            &                            & NMC~\cite{chen2021neural}  & 3.7               & \textit{94.7}            & 34.5              & 3.6                & 12.1 \\
                            &                            & DC~\cite{ju2002dual}       & \textit{3.5}               & 93.5            & 96.3             & \textit{3.4}                & \textit{11.6}               \\
                            &                            & NDC~\cite{chen2022neural}  & 3.6               & 94.3            & 43.1               & 3.5                & 11.9               \\ & &
                            FlexiCubes~\cite{FCubes} & 4.5 & 92.4 & 0   & 4.8 & 15.8 \\ &
                                                       & Ours                       & \textbf{3.3}      & \textbf{94.9}   & \textit{9.7}              & \textbf{3.2}       & \textbf{11.3}      \\ \hline
\multirow{6}{*}{Thingi10K}  & \multirow{6}{*}{Analytical} & MC~\cite{Lewiner2003}      & 5.1               & 61.4            & \textbf{0}        & 13.3               & 15.3               \\
                            &                            & NMC~\cite{chen2021neural}  & 4.1               & \textit{66.0}            & 32.4              & 11.9               & 12.2 \\
                            &                            & DC~\cite{ju2002dual}       & \textit{4.0}               & 63.8            & 105.6               & 11.5               & \textit{11.8}               \\
                            &                            & NDC~\cite{chen2022neural}  & \textit{4.0}               & 64.9            & 70.4              & \textit{11.3}               & 12.0               \\ 
                            & & FlexiCubes~\cite{FCubes} & 5.2 & 55.3 & 0   & 13.6 & 15.4 \\
                            &                            & Ours                       & \textbf{3.6}      & \textbf{68.0}   & \textit{15.7}              & \textbf{10.9}      & \textbf{11.6}       \\ \hline 
\multirow{5}{*}{Thingi10K}  & \multirow{5}{*}{Predicted~\cite{takikawa2021nglod}} & MC~\cite{Lewiner2003}      & 5.8               & 56.7            & \textbf{0} & 14.2               & 16.5 \\
                            &                            & NMC~\cite{chen2021neural}  & \textit{4.7}               & \textit{63.2}        &   30.1           &   \textit{12.6}        & \textit{12.9} \\
                            &                            & DC~\cite{ju2002dual}       & 6.2               & 54.0            & 230.4 & 14.8               & 15.0 \\
                            &                            & NDC~\cite{chen2022neural}  & \textit{4.7} & 62.8            & 42.6              & 12.8               & 13.1 \\
                            &                            & Ours                       & \textbf{4.2} & \textbf{65.7}   & \textit{15.9} & \textbf{12.2}      & \textbf{12.4} \\
\hline
\end{tabular}}
\caption{Quantitative mesh reconstruction results on the ABC and the Thingi10k datasets. The best scores are highlighted in bold and the second-best scores are in italics. The first two row-blocks correspond to input SDF which was computed analytically while in the last row-block, we mesh the SDF predicted by NGLOD~\cite{takikawa2021nglod}}.
\label{tab:combined_quantitative}
\end{table}

\subsection{Meshing analytical implicit functions}
We first consider the surface mesh reconstruction task given the \emph{ground truth} SDFs of shapes discretized on a regular grid. For this task, we train SDC on 3000 shapes from the $1^{st}$ split of the ABC dataset~\cite{Koch_2019_CVPR} for 100 epochs. We scale each shape to fit a unit sphere and estimate the SDF on a regular grid of dimension $64^3$ using SDFGen library~\footnote{\url{https://github.com/christopherbatty/SDFGen}}. We evaluate all methods on 300 shapes from the test set of the ABC dataset. Additionally, we evaluate on 300 shapes from the Thingi10K~\cite{T10k} dataset to demonstrate generalization to unseen data domains. We compare SDC against four baselines including two axiomatic baselines, namely the improved Marching Cubes (MC)~\cite{Lewiner2003} and the standard Dual Contouring (DC)~\cite{ju2002dual} where positions of each vertex are estimated by solving the Quadratic Error Function (QEF), and three of the recent Neural Meshing techniques, Neural Marching Cubes (NMC)~\cite{chen2021neural}, Neural Dual Contouring~\cite{chen2022neural} and FlexiCubes~\cite{FCubes} respectively. For fairness, we compare to variants of comparable network capacity and for FlexiCubes, we compare without test-time optimization.

We summarize our quantitative results in Table~\ref{tab:combined_quantitative}. Our self-supervised method SDC shows better performance and generalization to unseen data domains compared to the data-driven baselines across all metrics. While Marching Cubes~\cite{Lewiner2003} and FlexiCubes~\cite{FCubes} do not produce any self-intersecting faces, the vertex placement is restricted for these methods, leading to inferior results with respect to other reconstruction metrics. In the qualitative examples visualized in Figure~\ref{fig:NDCvsOurs}, we highlight an inherent issue with NDC and UNDC - they produce a considerable number of self-intersections along sharp edges. We reason this to be an undesirable trait that is inherited from the standard Dual Contouring as previously elaborated. SDC, on the other hand, bears a minimal number of self-intersections, as our training objective for mesh prediction is geometrically well-motivated. Additionally, we report quantitative and qualitative results for $128^3$ grid resolution in the Suppl.

\subsection{Meshing predicted implicit functions}
Differently from the previous section, where the SDF is computed analytically, here, we use an open-source implementation (with suggested hyperparameters) of a recent deep implicit network NGLOD~\cite{takikawa2021nglod} to predict SDF values on a regular grid. We evaluate the meshing efficacy of SDC over a set of 300 shapes from the Thingi10k~\cite{T10k} dataset. To avoid possible biases, we chose these 300 shapes to be different from the ones used in the previous section. Our quantitative results are summarized in Table~\ref{tab:combined_quantitative}. Consistent with our observation from the previous section, SDC outperforms all baselines. We provide a qualitative example of meshing a noisy SDF in Figure~\ref{fig:LearntSDF}. To generate the noisy SDF, we prematurely terminate NGLOD~\cite{takikawa2021nglod} fitting to the shape and evaluate different Neural Meshing methods. While all baselines produce a noisy reconstruction, SDC recovers a smoother surface while preserving sharp edges, thanks to our regularization and self-supervised losses. This suggests that our self-supervised training objectives and data augmentation are capable of generalizing to noisy imperfect predicted data \textit{without explicit training on this data}.

\subsection{Regularizing implicit surface learning}
As illustrated in the previous section, SDC successfully produces sharp meshes from both analytical and parameterized implicit functions. We now employ SDC for regularizing DIN training, focusing on the task of surface reconstruction using the ShapeNet dataset~\cite{shapenet2015}. In particular, we evaluate over 4 categories of objects, namely, cars, planes, tables, and cabinets with 1000 shapes per category as our training set, and set aside a separate set of 200 unseen shapes for evaluation. We compare our regularization against three possible baselines. Firstly, we train Curriculum DeepSDF (CSDF)~\cite{CSDF} over each category separately. Secondly, we re-train the same network by enforcing the Eikonal constraint, \ie enforcing a unit norm for SDF's gradient in a setup similar to Implicit Geometric Regularization~\cite{IGR}. Thirdly, we use the supervised Dual Contouring baseline NDC~\cite{chen2022neural} as the mesh-based regularizer. More specifically, we replace SDC with NDC in computing the regularization introduced in \eq{sdr}. For each shape, we sample 400,000 points in the shape volume, aggressively near the surface following~\cite{park2019deepsdf} to supervise the SDF prediction. For fairness, we use the same sampled points for all methods we compare.

\begin{figure}
\begin{center}
  \includegraphics[width=\linewidth]{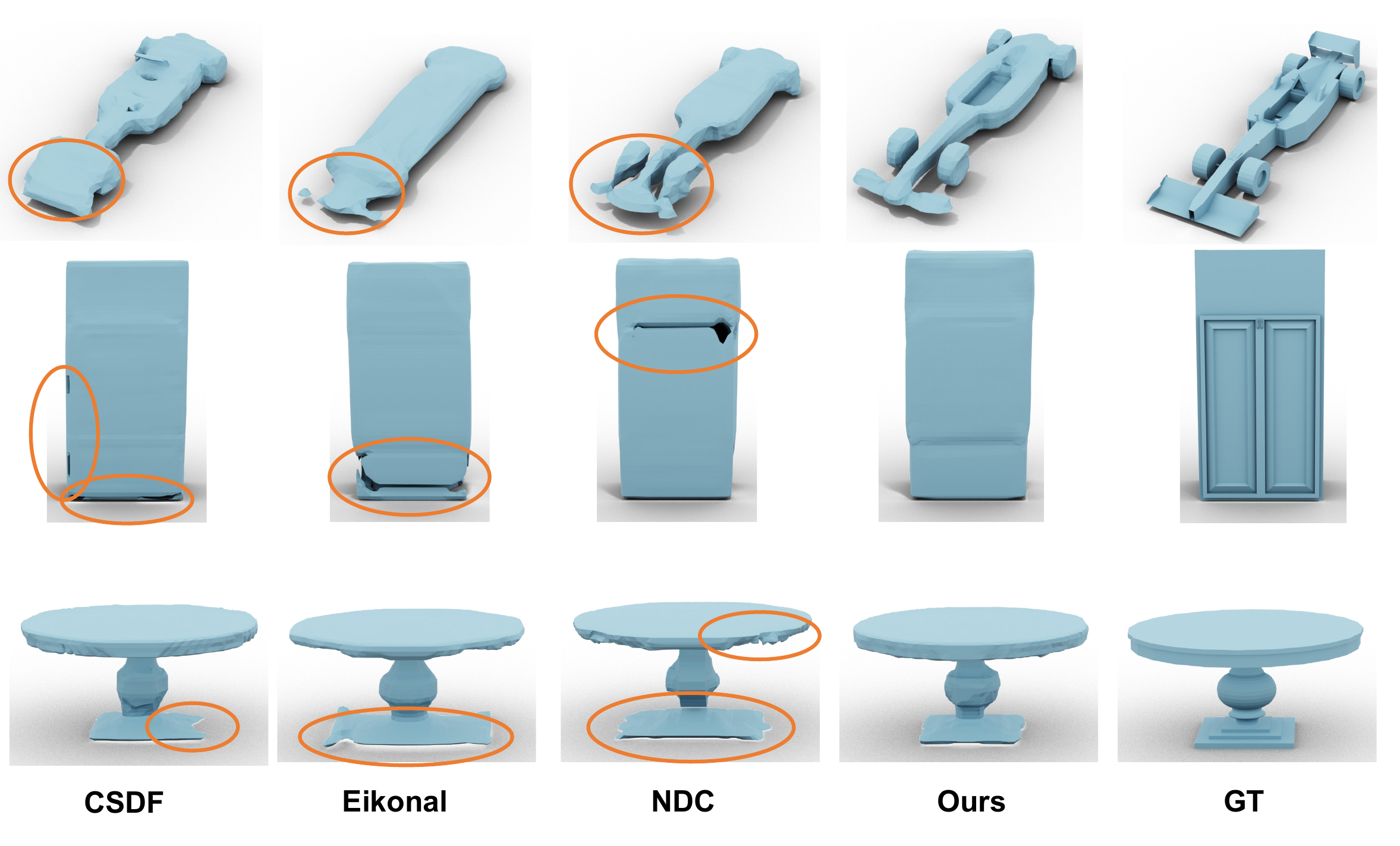}
  \caption{Qualitative examples comparing surface reconstruction from the ShapeNet~\cite{shapenet2015} dataset. The columns compare different methods, whose training objectives differ. Our mesh-based regularization produces reasonable results compared to baselines.\vspace{-1mm}}
\label{fig:regul}
\end{center}
\end{figure}

We summarize our quantitative results in Table~\ref{tab:DINRegul}. We observe that regularizing the network using SDC produces a noticeable improvement in reconstruction across all metrics in comparison to baselines. Moreover, we observe a poorer reconstruction when using NDC~\cite{chen2022neural} for regularization. We argue that this is due to their inability to handle noisy predicted SDF values as inputs as this approach learns to emulate solution to QEF (c.f. \eq{QEF}), which is ill-defined for imperfect implicit functions. Our self-supervised loss functions do not rely on the ill-defined vertices produced by QEF minimization and align produced meshes to the input SDF grids, resulting in more plausible surface predictions. We believe this difference to be the key reason behind SDC's efficacy as a regularizer for training DINs. We also show three qualitative results in Figure~\ref{fig:regul}. In some examples, we also observe topological differences although our regularization does not explicitly penalize it. Since the loss function which we minimize is highly non-convex, we believe that our regularization has aided in discovering more ``plausible'' latent space which could lead to better quality of implicit surfaces. We report additional qualitative and quantitative results in Suppl demonstrating the regularizer's generalization capabilities with other DINs.

\begin{table}[]
\centering
\resizebox{0.9\columnwidth}{!}{%
\begin{tabular}{@{}lcccc@{}}
\toprule
\rowcolor{gray!25}
Method     & CD ($\downarrow$) & NC ($\uparrow$) & 3D IoU ($\uparrow$) & LSD-A ($\downarrow$) \\ \midrule
CSDF~\cite{CSDF}      & 3.0 & 78.1 & 83.9 &  3.1\\
+ Eikonal~\cite{IGR}  & 3.2 & 78.0 & 83.0 &  3.2\\
+ NDC~\cite{chen2022neural}      & \textit{2.9} & \textit{78.4} & \textit{84.8} &  \textit{2.9}\\
+ SDC      & \textbf{2.6} & \textbf{78.9} & \textbf{86.0} & \textbf{2.7} \\
\bottomrule
\end{tabular}%
}
\caption{Comparison of surface reconstruction accuracy across object categories from the ShapeNet dataset~\cite{shapenet2015}. Methods are trained with different regularization (see text) while evaluated alike.\vspace{-2mm}}
\label{tab:DINRegul}
\end{table}

\subsection{Single view reconstruction}
In this section, we consider the task of reconstructing surface meshes from images. We use 4 categories from the ShapeNet dataset~\cite{shapenet2015}, namely planes, chairs, rifles, and tables. We train and evaluate our approach and baseline (using the official codebase) on the same training and evaluation split for a fair comparison. We compare against three baselines, namely,  MeshSDF~\cite{remelli2020meshsdf}, DISN~\cite{DISN}, and a variant of our approach referred to as WoBW. More specifically, WoBW meshes the implicit function produced by pre-trained SV-DIN using SDC, without the joint fine-tuning. For MeshSDF~\cite{remelli2020meshsdf} and DISN~\cite{DISN} we use the official codebase for re-training and evaluation.

\begin{figure}
\begin{center}
  \includegraphics[width=\linewidth]{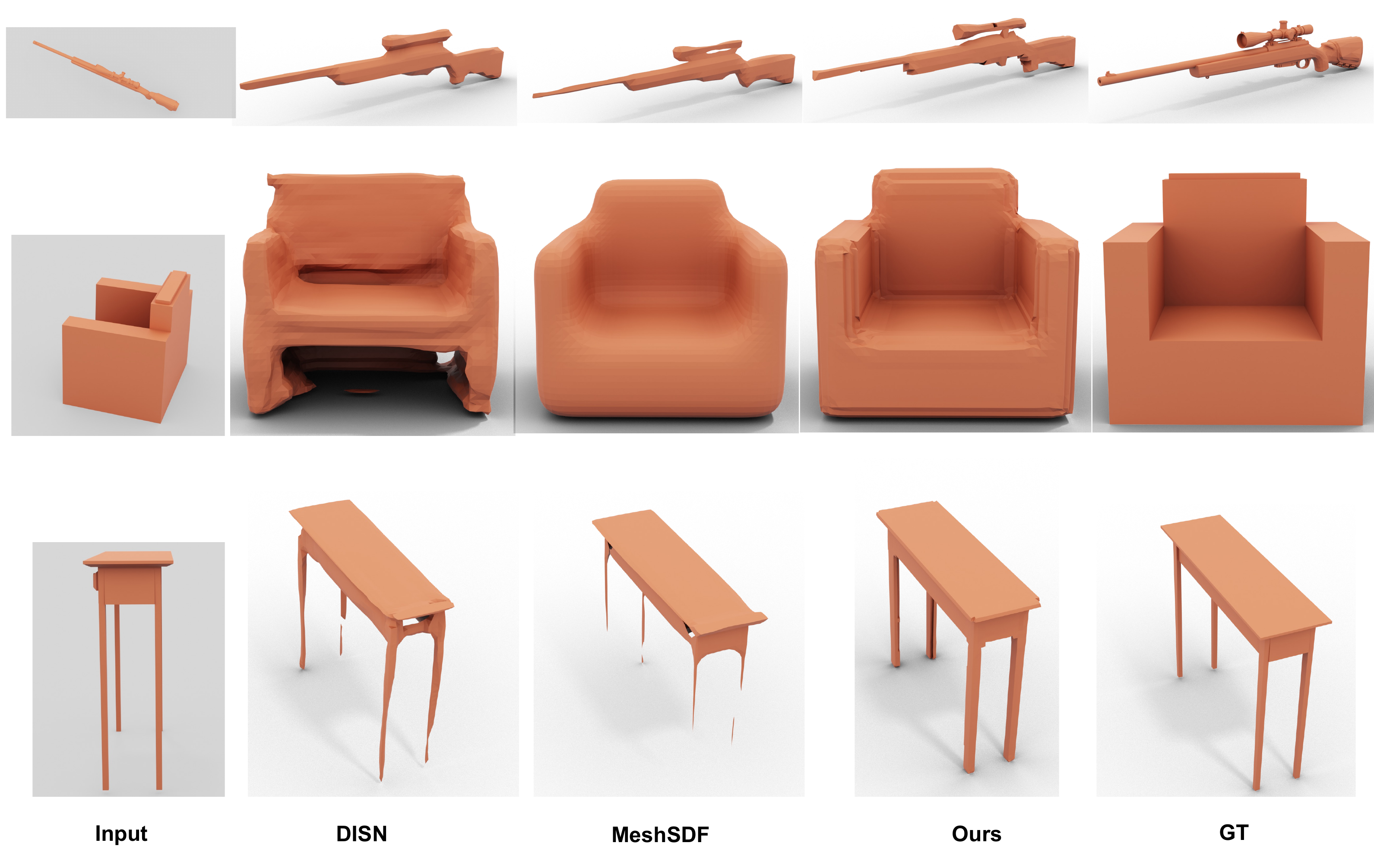}
  \caption{Qualitative comparison of Single View Reconstruction (SVR) accuracy among three objects from ShapeNet~\cite{shapenet2015} dataset. Input denotes the input view and GT denotes ground-truth mesh. SDC produces more sharper and plausible surfaces in comparison to baselines.}
\label{fig:SVRQual}
\end{center}
\end{figure}

Quantitative results averaged across 4 object categories are summarized in Table~\ref{tab:fromIM}. Notably, our SDC outperforms MeshSDF~\cite{remelli2020meshsdf} without additional test-time optimization fitting a mesh to the rendering. In addition, our end-to-end model consistently outperforms the WoBW baseline, implying the efficacy of our mesh-based regularization and end-to-end training. We also show three qualitative examples in Figure~\ref{fig:SVRQual}. Our SDC demonstrates an improved ability to reconstruct sharp features in the predicted meshes, resulting in surfaces more faithful to the ground truth. In the third row, we highlight an example showing significant improvements in the quality of predicted geometry. It shows that our method is capable of better thin surface reconstructions for the same inputs.

\begin{table}[]
\resizebox{0.9\columnwidth}{!}{%
\begin{tabular}{@{}lcccc@{}}
\toprule
\rowcolor{gray!25}
Method                          & CD ($\downarrow$) & NC ($\uparrow$) & SSIM ($\uparrow$) & LSD-A ($\downarrow$) \\ \midrule
DISN~\cite{DISN}                & 13.0 & 70.2 & 82.0 & 10.5\\
MeshSDF~\cite{remelli2020meshsdf} & 11.3 & 71.4 & 84.1 & 9.1\\
WoBW                            & \textit{9.9}  & \textit{72.0} & \textit{86.4} & \textit{8.1}\\
Ours                            & \textbf{9.1}  & \textbf{72.8} & \textbf{88.5} & \textbf{7.7}\\
\bottomrule
\end{tabular}%
}
\caption{Quantitative results for single-view reconstruction across different object categories from ShapeNet~\cite{shapenet2015} dataset.}
\label{tab:fromIM}
\end{table}

\section{Conclusion and future work}
We introduced SDC, a Self-Supervised training approach for Neural Dual Contouring. Differently from previous work, we do not fit the generated mesh to an axiomaticly produced proxy mesh but instead use geometrically motivated self-supervised losses which only depends on the input SDF. SDC shows consistent improvements over baselines when applied to end-to-end surface reconstruction and meshing tasks. In addition, our work forges a link between signed distance fields and meshes and shows how the former can be regularized by guiding the network to produce distance fields better corresponding to resulting meshes.
Our meshing and regularization are applied on a regular grid, which limits the resolution of a reconstructed mesh. It would be interesting to explore adaptive grids within the learning framework. Finally, another related and interesting scope for future work is the exploration of strict manifoldness conditions~\cite{ManifoldDC} and theoretical guarantees of self-intersection free surfaces.

{\paragraph{Acknowledgements }
This work was granted access to the HPC resources of IDRIS under the allocation 2023-AD011013104R2 made by GENCI. Parts of this work were supported by the ERC Starting Grant 758800 (EXPROTEA), ERC Consolidator Grant 101087347 (VEGA), ANR AI Chair AIGRETTE, and gifts from Ansys and Adobe Research.}

\setcounter{page}{1}
\maketitlesupplementary

\section{Introduction}
\label{sec:rationale}

This document serves as the supplemental material to our main work ``Self Supervised Dual Contouring''. We perform ablation studies on potential alternative reconstruction losses in Section~\ref{sec:abl}. Then, we provide additional implementation details pertaining to our three experimental settings in Section~\ref{sec:impldet}. We elaborate on various evaluation metrics used in the main paper in Section~\ref{sec:evalM}. Finally, we provide additional quantitative evaluation and qualitative results in Section~\ref{sec:qualt}. Our entire code will be released upon publication.

\section{Ablation Studies}
\label{sec:abl}
We perform Ablation studies over our self-supervised loss function and sampling strategy for applying our proposed mesh-based regularization.

\subsection{Reconstruction losses}

In this section, we compare alternative loss functions for meshing a given implicit function. To recall, we refer to our loss as Self-Supervised as our learning framework does not rely on a reference object and our two loss functions are computed purely in terms of input grid of SDF. Herewith, we compare loss functions which uses explicit supervision w.r.t. a reference mesh to our self-supervised loss function. \Ie, instead of aligning the generated surface to best fit the input SDF, we compare our SDC with loss functions that try best aligns the generated surface to some discrete set of surface samples. These discrete surface samples are points sampled on the mesh from ABC~\cite{Koch_2019_CVPR} dataset, which we used for training SDC as stated in our main paper.  We remove our SDC losses and replace them with different loss functions as elaborated below:
\begin{enumerate}
    \item \textbf{Random Sampling}: We randomly sample points on generated surface and ground truth surface and minimize the Chamfer's Distance between them. 

    \item \textbf{Area Sampling}: We use an area based sampling where we sample the generated surface proportional to the area of triangles. Then, Chamfer's Distance is minimized between the aforementioned samples and ground-truth surface samples. 

    \item \textbf{Vert CD}: We apply Chamfer Distance, between generated vertex (produced by SDC) and mean coordinate of point samples within the same voxel as the generated vertex. This loss function is local with locality defined by voxel cell. 

    \item \textbf{Second-Order CD}: Minimizing the distance between two surfaces has been well-studied in the context of shape registration with theoretical guarantees~\cite{LocRegSGP,Pottmann2004,Pottmann2003}. To that end, we consider quadratic approximation of point-point distance proposed in \cite{LocRegSGP} as our baseline. In particular, \cite{LocRegSGP} uses local curvature information of the surface to incorporate second order information into the function which measures the distance between query point and reference surface. In our case, we consider the query point to be the mesh vertex predicted by our network $h_\phi$ and the reference surface to be the ground truth mesh from our training dataset, ABC~\cite{Koch_2019_CVPR}. Since we explicitly use the mesh, we consider this to be supervised baseline. This supervised training objective is given as follows:
    \begin{equation*}
    \begin{aligned}
    \mathcal{L} = & \hat{\delta}_1\left(\vec{e}_1 \cdot(\mathbf{x}-\mathbf{y})\right)^2+\hat{\delta}_2\left(\vec{e}_2 \cdot(\mathbf{x}-\mathbf{y})\right)^2+ \\
    & (\vec{n} \cdot(\mathbf{x}-\mathbf{y}))^2,
    \end{aligned}
    \end{equation*}
    where $\mathbf{x}$ denotes the query point, $\mathbf{y}$ denotes the closest point on the surface where the surface normal is given by $\vec{n}$ and the direction of principal curvature is given by $\vec{e_1},\vec{e}_2$. Finally, $\delta_1$, $\delta_2$ denote the magnitude of principal curvatures at $\mathbf{y}$.

    \item \textbf{W/o NC}: We do not use the normal consistency loss and only use $\mathcal{L}_D$ defined in Eqn.3 of the main paper.

    \item \textbf{Ours}: Denotes the loss function which we report in the paper. 
\end{enumerate}

Our quantitative results are summarized in Table~\ref{tab:Ablation}. We observe a noticeable improvement in performance when using our self-supervised loss function compared to supervision with surface sampling. As discussed in related works~\cite{tatarchenko19cvpr,wu21neurips} sampling surfaces with discrete points could lead to attraction of points to a single source (or sink) at regions of uneven sampling density. This could potentially explain the higher self-intersection. Also, more importantly, Self-Supervised loss (Ours) shows a significant improvement over supervised baseline (Second-Order). The reported experiments were performed on the test-set of the ABC dataset defined in the main paper.

\begin{table}[]
\resizebox{\columnwidth}{!}{%
\begin{tabular}{cllllll}
\hline
\multicolumn{1}{l}{Type} & Method & CD ($\downarrow$) & NC ($\uparrow$) & SI ($\downarrow$) & ECD ($\downarrow$) & LSD-P ($\downarrow$) \\ \hline
\multirow{4}{*}{\begin{tabular}[c]{@{}c@{}}Explicit \\ Supervision\end{tabular}} & Random Sampling & 3.90 & 90.20 & 77.60 & 3.85 & 13.1 \\ 
 & Area Sampling & 3.72 & 91.60 & 50.44 & 3.70 & 12.6 \\ 
 & Vert CD & 3.52 & 91.70 & 72.70 & 3.54 & 11.9 \\ 
 & $2^{nd}$ Order CD & 3.47 & 91.00 & 14.55 & 3.49 & 11.8 \\ \hline
\multirow{2}{*}{\begin{tabular}[c]{@{}c@{}}Implicit \\ Supervision\end{tabular}} & W/o NC & 3.40 & 93.80 & 11.64 & 3.40 & \textbf{11.3} \\ 
 & Ours & \textbf{3.30} & \textbf{94.90} & \textbf{9.67} & \textbf{3.20} & \textbf{11.3} \\ \hline
\end{tabular}%
}
\caption{Ablation study on different unsupervised losses for the task of meshing an implicit function. }
\label{tab:Ablation}
\end{table}

\subsection{Sampling for regularization}

\begin{figure*}
\begin{center}
\includegraphics[width=\textwidth]{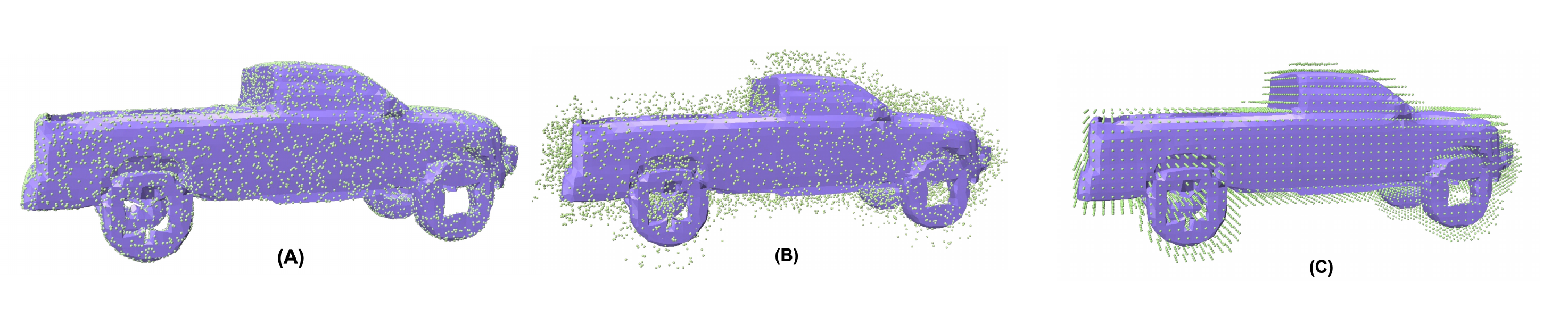}
\end{center}
\caption{Depicting point samples (in green) at which our signed distance based regularization is applied. The implicit function is meshed using SDC and rendered in purple. (A) denotes surface sampling, (B) Denotes irregular near-surface sampling and (C) is regular near-surface sampling.}
\label{fig:SamplingEffect}
\end{figure*}

\begin{table}[]
\centering
\resizebox{\columnwidth}{!}{%
\begin{tabular}{@{}lccccl@{}}
\toprule
Sampling Type & \begin{tabular}[c]{@{}l@{}}CD $\downarrow$\\ (x10\textasciicircum{}3)\end{tabular} & \begin{tabular}[c]{@{}l@{}}NC  $\uparrow$\\ (\%)\end{tabular} & \begin{tabular}[c]{@{}l@{}} IoU  $\uparrow$\\ (\%)\end{tabular}  &  \begin{tabular}[c]{@{}l@{}}LSD-A  $\downarrow$\\ (x10\textasciicircum{}3)\end{tabular} & \begin{tabular}[c]{@{}l@{}}Time  $\downarrow$\\ (m-sec)\end{tabular}                    \\ \midrule
Surface       & 3.6 & 74.2 & 84.9 & 2.8 & 2.3                     \\
Volume        & 2.8 & 77.5 & 86.9 & 2.8 & 2.3 \\
Reg Grid      & \textbf{2.6} & \textbf{79.0} & \textbf{87.3} & \textbf{2.7} & \textbf{2.1} \\ \bottomrule
\end{tabular}%
}
\caption{Comparing quantitative reconstruction results and timing between different sampling strategies for applying our mesh-based regularization. }
\label{tab:SDRAbalation}
\end{table}

In this section, we ablate various sampling strategies which could be used to establish the points for which $\mathcal{L}_\mathrm{SDR}$ (c.f Eqn.7, main paper) can be computed. In particular, we compare between three types of sampling points in space for which SDF prediction is regularized w.r.t. the mesh produced by SDC. Firstly, we compare between points that are defined on a regular grid, close to the surface of the mesh. Secondly, we consider points sampled uniformly on the surface of the mesh produced by SDC. Thirdly, we add small random displacement along the normal vector such that the points on the mesh are close to but not necessarily on the surface of the mesh. The three sampling strategies mentioned above are visualized in Figure~\ref{fig:SamplingEffect}. We sample 20,000 points for each strategy. We compare the reconstruction accuracy of the implicit surface while using the proposed regularization along with different sampling strategies. The quantitative results are summarized in Table~\ref{tab:SDRAbalation}. We observe that a regular sampling along the grid shows overall better performance compared to random sampling. We also report the average time for performing a single forward pass using the aforementioned regularization. We observe that using a regular sampling strategy is less time-consuming in comparison to other sampling strategies. This is because, we can \emph{re-use} the SDF values computed at those grid points as SDC requires SDF values at grid points to reconstruct a mesh. This is unlike the latter two cases where another forward pass through CSDF~\cite{CSDF} is required to determine the SDF values. All experiments were performed on Ampere-A100 GPUs for fairness in comparison. 

\section{Additional implementation details}
\label{sec:impldet}
We first provide additional details on estimating normals and then provide implementational details for the three experiments performed in our main paper.

\subsection{Meshing implicit surfaces}

\begin{figure}
\begin{center}
\includegraphics[width=\linewidth]{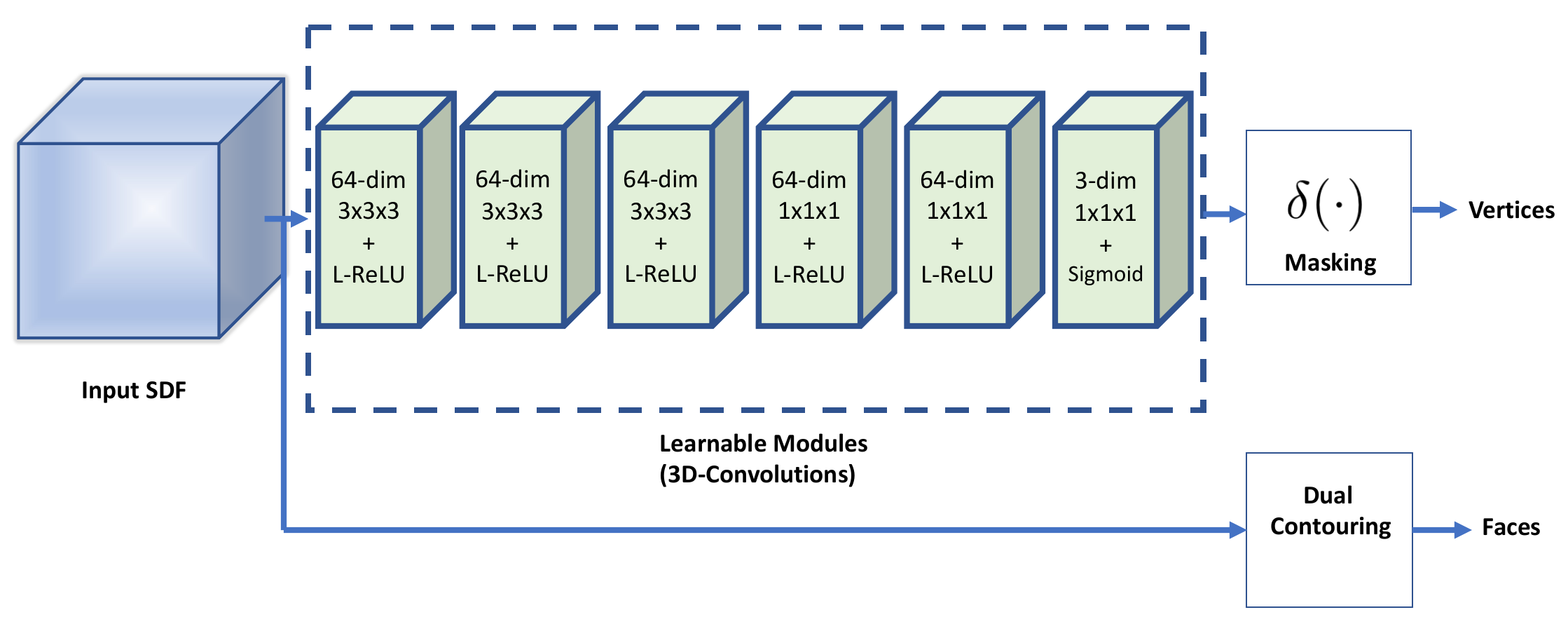}
\end{center}
\caption{Detailed depiction of SDC framework. Starting from input SDF defined on a regular grid, we predict the vertices of the mesh while constructing the faces following the Dual Contouring~\cite{ju2002dual} technique.}
\label{fig:DetailedArch}
\end{figure}

\paragraph{Normal Estimation.} We estimate the normals from grid of SDF using Finite-Difference gradient estimation. Given a 3D scalar field (SDF in this case) \( f(x, y, z) \), where \( x, y, z \) are grid coordinates, we can compute the gradient at each grid point using finite differences.

For an interior grid point, we use the 5-point stencil:
\small
\begin{align*}
\begin{split}
\frac{\partial f}{\partial x} &= \frac{-f(x+2h, y, z) + 8f(x+h, y, z)}{12h} \\
&\quad - \frac{8f(x-h, y, z) - f(x-2h, y, z)}{12h},
\end{split}
\\
\begin{split}
\frac{\partial f}{\partial y} &= \frac{-f(x, y+2h, z) + 8f(x, y+h, z)}{12h} \\
&\quad - \frac{8f(x, y-h, z) - f(x, y-2h, z)}{12h},
\end{split}
\\
\begin{split}
\frac{\partial f}{\partial z} &= \frac{-f(x, y, z+2h) + 8f(x, y, z+h)}{12h} \\
&\quad - \frac{8f(x, y, z-h) - f(x, y, z-2h)}{12h}.
\end{split}
\end{align*}
\normalsize

For boundary grid points, using a 2-point stencil:
\begin{align*}
\text{At the start boundary:} \\
\frac{\partial f}{\partial x} &= \frac{f(x+h, y, z) - f(x, y, z)}{h}, \\
\frac{\partial f}{\partial y} &= \frac{f(x, y+h, z) - f(x, y, z)}{h}, \\
\frac{\partial f}{\partial z} &= \frac{f(x, y, z+h) - f(x, y, z)}{h}. \\
\text{At the end boundary:} \\
\frac{\partial f}{\partial x} &= \frac{f(x, y, z) - f(x-h, y, z)}{h}, \\
\frac{\partial f}{\partial y} &= \frac{f(x, y, z) - f(x, y-h, z)}{h}, \\
\frac{\partial f}{\partial z} &= \frac{f(x, y, z) - f(x, y, z-h)}{h}.
\end{align*}
The resulting gradient vector at any grid point is then given by:
\[
\mathbf{df} = \left( \frac{\partial f}{\partial x}, \frac{\partial f}{\partial y}, \frac{\partial f}{\partial z} \right).
\]
To get the normal vector, normalize the gradient:
\[
\mathbf{n} = \frac{\mathbf{df}}{||\mathbf{df}||}.
\]

\paragraph{Training Details.} We train our SDC with ADAM optimizer~\cite{Kingma2014AdamAM} for 200 epochs with a learning rate of $0.0001$. The final layer of SDC is a scaled-sigmoid, with scaling factor of $\sigma=10$ to facilitate learning. We use sigmoid activation so that the predicted vertex does not leave the cell. For all experiments in our main paper, we train on 3,000 shapes from the ABC dataset. All shapes are scaled to fit a unit-sphere. We pre-compute SDFs of all clean shapes at $64^3$ resolution and apply augmentation on-the fly. We used $\alpha_1=\alpha_2=\alpha_3=\alpha_4=0.01$ in all our experiments. Detailed architecture depicting SDC is illustrated in Figure~\ref{fig:DetailedArch}. 

\paragraph{Why avoid supervision?} The supervised data-driven approach NDC~\cite{chen2022neural} emulates standard Dual Contouring using a local augmentation strategy for individual grid-cells, mitigating noisy triangulation from DC (refer to Figure 5 in ~\cite{chen2022neural}). Though this augmentation effectively enhances triangle quality, it remains an ad-hoc and heuristic solution. Conversely, SDC determines vertex positioning through a loss function mirroring the motivation of standard Dual Contouring, without relying on ad-hocs. Instead of addressing vertex positioning as a potentially ill-posed Linear Least Squares problem, we employ iterative optimization, a fitting choice for training neural networks. The axiomatic method's supervision is limiting since training is feasible only with pristine input data. SDC neither relies on clean input data nor a heuristic augmentation to circumvent poor triangulation. 

\paragraph{Why less self-intersection?}
As mentioned above, NDC~\cite{chen2022neural} is trained to emulate standard Dual Contouring vertices. Since Dual Contouring computes the intersection of the surface inside each voxel cube using a linear interpolation scheme, this implies, DC assumes the surface to be a smooth function that can be approximated by a linear function. However, at sharp edges and corners, the surface is not smooth, and this linear interpolation produces self-intersecting faces. On the other hand, the zero-level set of any function is free of self-intersection. Therefore, since SDC produces vertices to minimize discrepancy between SDFs, it is geometrically better motivated and as a result produces fewer self-intersections.

\paragraph{Timings.}
\begin{table}[]
\centering
\resizebox{0.6\columnwidth}{!}{%
\begin{tabular}{@{}llll@{}}
\toprule
\multicolumn{1}{c}{Method} & Pre-Process & Training & Inference \\ \midrule
NDC & 9.0 & \textbf{0.07} & \textbf{0.025} \\
UNDC & 9.0 & 0.14 & 0.05 \\
SDC & \textbf{2.7e-3} & 0.10 & \textbf{0.025} \\ \bottomrule
\end{tabular}%
}
\caption{We compare timing between our meshing method (SDC) and two supervised Dual Contouring based baselines, NDC and UNDC. Reported timings are in seconds per-shape for $64^3$ grid.  }
\label{tab:timing}
\end{table}
We report pre-processing, training and inference timing of our SDC and compare against supervised baselines NDC and UNDC~\cite{chen2022neural} in Table~\ref{tab:timing}. SDC has a negligible pre-processing time as it only involves scaling of the mesh. On the other-hand, NDC and UNDC computes dual-contouring ground-truth vertices which involves solving a linear system of equation for every occupied voxel and hence their pre-processing time is costlier. However, since our approach relies on point-to-face distance estimation (c.f. Eqn.7, main paper), our training is slightly costlier than both NDC and UNDC. On the other-hand, UNDC requires twice the training effort as it separately learns connectivity information. In practise, we require roughly 10hrs for training SDC. Similarly, for inference, UNDC requires twice the amount of time per-shape while SDC and NDC have similar inference time. All experiments were performed on Ampere-A100 GPU on a grid of resolution $64^3$, averaged across our training set. 

\subsection{Regularizing DIN}
As mentioned in the main paper, we used Curriculum DeepSDF~\cite{park2019deepsdf} as our choice of DIN and followed the same hyper-parameters and training strategy advocated by the respective author. Since the curriculum learning strategy increases the level-of-detail of the reconstructed SDF gradually, we apply our regularization for the last 200 epochs. We use the weighting factor of our regularization term $\epsilon=10$. We used points sampled on a regular grid that are closer to the surface as illustrated in Figure~\ref{fig:SamplingEffect}. To construct a mesh by SDC, we used a regular $64^3$ grid. We trained all baselines, including ours for a total of 2000 epochs. 

\subsection{Mesh from Image(s)}
\begin{figure}[h]
\begin{center}
\includegraphics[width=\linewidth]{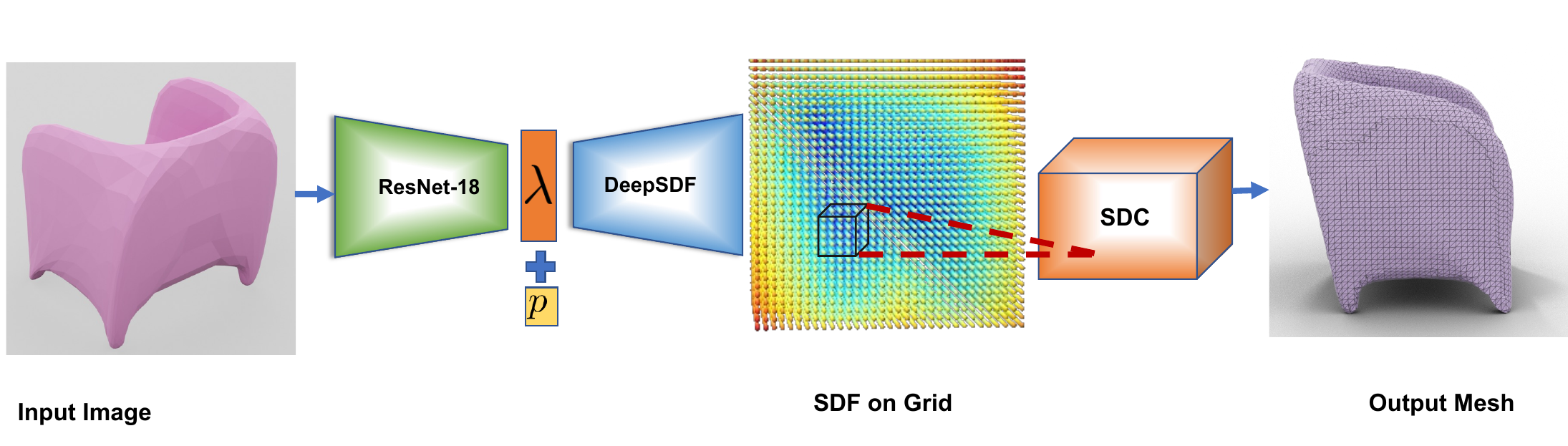}
\end{center}
\caption{Network architecture for joint training for the task of Single View Reconstruction. Starting from an image, we construct an implicit representation via a latent code and then reconstruct the mesh using SDC.}
\label{fig:E2E}
\end{figure}

We jointly train a ResNet-18~\cite{Resnet} and DeepSDF~\cite{park2019deepsdf} to learn an implicit surface for each image from the training set. We train over ShapeNet~\cite{shapenet2015} dataset and use the rendering provided by~\cite{DISN} as the input to ResNet-18 encoder. Rendering of each shape is encoded into a latent vector using ResNet-18, which is then concatenated along with a query point for which SDF is learnt by DeepSDF. To obtain query points, we sample 400,000 points close to the surface similar to~\cite{park2019deepsdf}. We train for a total of 2,000 epoch per-category using ADAM~\cite{Kingma2014AdamAM} optimizer. For the baselines, use same hyper-parameters provided by the author for a fair comparison. We visualize the network used for end-to-end training in Figure~\ref{fig:E2E}.

\section{Evaluation Metrics}
\label{sec:evalM}
We provide more details on various evaluation metrics used throughout the paper.:
\begin{itemize}
    \item Chamfer Distance (CD) calculates a symmetric distance between two sets of points. Given $\chi_1$ and $\chi_2$ to be two point clouds, 
    \begin{equation*}
    \begin{split}
        CD(\chi_1, \chi_2) &= \frac1{|\chi_1|}\sum_{x \in \chi_1}\min_{y \in \chi_2}||x - y||_2 \\
        & + \frac1{|\chi_2|}\sum_{y \in \chi_2}\min_{x \in \chi_1}||y - x||_2.
    \end{split}
    \end{equation*}
    \item The Normal Consistency (NC) is often used as a metric for 3D surface reconstruction tasks to measure how well the estimated surface is consistent with the underlying geometry of the object:
    \begin{equation*}
        \mathrm{NC} = \frac{1}{N} \sum_{i=1}^N (1 - \cos(N_i, N^*_i)) \times 100\%, 
    \end{equation*}
    $N_i$ and $N^*_i$ are the estimated and ground truth surface normals, respectively, for the $i^{th}$ point on the surface, and $\cos(N_i, N^*_i)$ represents the cosine similarity between the two vectors. The normalization term $\frac{1}{N}$ ensures that the metric is independent of the number of points on the surface.
    
    \item Edge Chamfer distance (ECD) is similar to the regular Chamfer Distance, but it is calculated for two sets of points, sampled on the edges of considered meshes. This metric better gauges how well the edges are reconstructed, or, another way of measuring sharpness. 
    
    \item 3D Intersection-over-Union (IoU) is a metric used to compare pairs of 3D shapes, represented as 3D voxel grids $G^1, G^2$. It considers a ratio of the number of occupied voxels in the intersection of two occupancy grids to the number of occupied voxels in the union of occupancy grids:
    \begin{equation*}
        IoU(G^1, G^2) = 100 * \frac{\sum_{ijk} G^1_{ijk} \land G^2_{ijk}}{\sum_{ijk} G^1_{ijk} \lor G^2_{ijk}},
    \end{equation*}
    where, $i,j,k$ are indices of voxel-cell along x,y,z dimensions. 

    \item Precise Level Set Discrepancy (LSD-P). We propose this metric to gauge the discrepancy in the signed distance values between a reconstructed mesh and the ground truth mesh at fixed points in space. The fixed points are sampled on a regular grid $\mathcal{G}$ that are close to the surface of the ground truth mesh. This metric is defined as follows:
    \begin{equation*}
        \text{LSD-P} = \sum_{g \in \bar{\mathcal{G}}} || \mathrm{abs}(\mathbf{d}^{p}(\mathcal{M^*}, g)) -  \mathrm{abs}(\mathbf{d}^{p}(\mathcal{M}, g)) ||_2,
    \end{equation*}
    where $\bar{\mathcal{G}}$ denotes the grid points sampled close to the ground truth surface $\mathcal{M}^{*}$. Following the similar definition in our main paper, $\mathbf{d}^{p}$ denotes the distance between g and its closest point on a given surface. $\mathcal{M}$ refers to the reconstructed surface.

    \item Approximate Level Set Discrepancy (LSD-A). We introduce this metric to gauge the discrepancy in zero-level set between generated mesh and the ground truth mesh in circumstances where ground truth mesh might not be water-tight. This metric was used in our main paper for experiments pertaining to ShapeNet~\cite{shapenet2015} dataset since it contains meshes that are non-watertight and estimating analytic SDF is ill-defined. We first sample points on the ground truth mesh and then measure the point-to-face distance to the generated mesh's closest face. It is defined as follows:
    \begin{equation*}
        \text{LSD-A} = \frac{1}{N} \sum_{q \in \mathcal{M}^*} d^p(q, \mathcal{M}).
    \end{equation*}

    \item SSIM (Structural Similarity Index) is a quality metric that measures the similarity between two images. The metric measures three components of image similarity: luminance, contrast, and structure. The structural information of two images include features such as edges, contrast, and texture. 

    \item Self-Intersection. We use VCGlib~\footnote{\url{https://github.com/cnr-isti-vclab/vcglib/}} to determine face self-intersections. Initially, the algorithm employs spatial indexing using a grid to efficiently organize the mesh faces. This structure allows for identification of potentially intersecting faces by comparing their bounding boxes, thereby significantly reducing the number of detailed intersection tests required. Once potential intersecting pairs are identified, the algorithm determines actual geometric intersections. It first assesses the number of shared vertices between each pair of faces. If no vertices are shared, a direct triangle-to-triangle intersection test is performed. For pairs with a single shared vertex, the algorithm evaluates by creating segments from the non-shared vertices of each face, offset towards the shared vertex, and then examines these segments for intersections with the opposite triangle. For pairs of triangles that share an edge, intersection test is performed by checking the position of the third vertex. 

\end{itemize}

\section{Additional Results}
\label{sec:qualt}

\begin{table}[h]
\centering
\resizebox{0.9\columnwidth}{!}{%
\begin{tabular}{cllllll}
\hline
\rowcolor{gray!25}
\multicolumn{1}{l}{Dataset} & Method                     & CD ($\downarrow$) & NC ($\uparrow$) & SI ($\downarrow$) & ECD ($\downarrow$) & LSD-P ($\downarrow$) \\ \hline
\multirow{5}{*}{ABC}        
                            & MC~\cite{Lewiner2003}      & 3.62               & 93.69            & \textbf{0}                  & 2.72                & 9.60               \\
                            & NMC~\cite{chen2021neural}  & 3.53               & 96.18            & 12.00              & 2.16                & 9.02 \\
                            & NDC~\cite{chen2022neural}  & 3.39               & 95.70            & 14.61              & 2.22                & 9.26               \\
                            & Ours                       & \textbf{3.15}               & \textbf{96.34}            & 6.84               & \textbf{2.04}                & \textbf{8.60}      \\ \hline
\multirow{5}{*}{Thingi10K}  
                            & MC~\cite{Lewiner2003}      & 3.89               & 64.10             & \textbf{0}                & 11.70               & 10.98               \\
                            & NMC~\cite{chen2021neural}  & 3.58               & 68.34            & 27.40              & 10.92               & 10.56 \\
                            & NDC~\cite{chen2022neural}  & 3.52               & 67.20            & 40.82              & 11.08               & 10.60               \\
                            & Ours                      & \textbf{3.41}               & \textbf{69.8}            & 14.80              & \textbf{10.21}               & \textbf{9.96}       \\ \hline 
\hline
\end{tabular}
}
\caption{Quantitative mesh reconstruction results on the ABC and the Thingi10k dataset evaluated on a $128^3$ SDF grid.}
\label{tab:quant128}
\end{table}

\begin{figure*}
\includegraphics[width=1\linewidth]{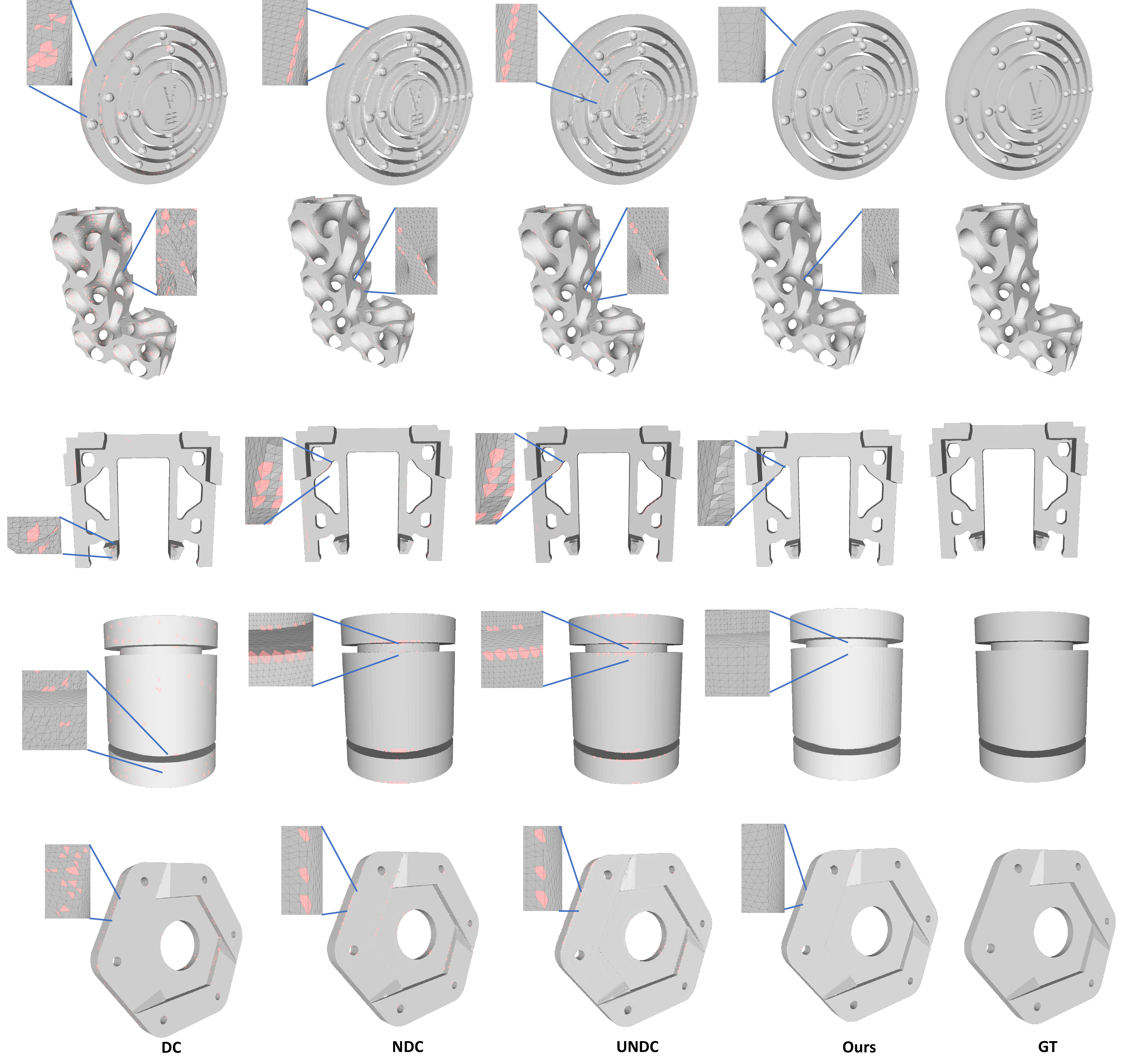}
\caption{Qualitative comparison between meshes reconstructed from an input SDF of $128^3$ resolution between DC~\cite{ju2002dual}, NDC and UNDC~\cite{chen2022neural}. Self-intersecting faces are highlighted in red. Please zoom-in to see the detailed tessellation. First three rows are from Thingi10K dataset and last two rows are from the ABC dataset.}  
\label{fig:MeshingQual}
\end{figure*}

In this section, we provide additional qualitative and quantiative results. Throughout our main paper, we considered regular grids of size $64^3$ for all our experiments. Now, we show that our method can be scaled to grid resolution of $128^3$ without any additional training. SDC still produces minimal self-intersection and superior reconstruction in comparison to supervised baselines. We summarize our quantiative results in Table~\ref{tab:quant128} and provide qualitative illustration in Figure~\ref{fig:MeshingQual}. 

\begin{figure*}
\includegraphics[width=0.9\linewidth]{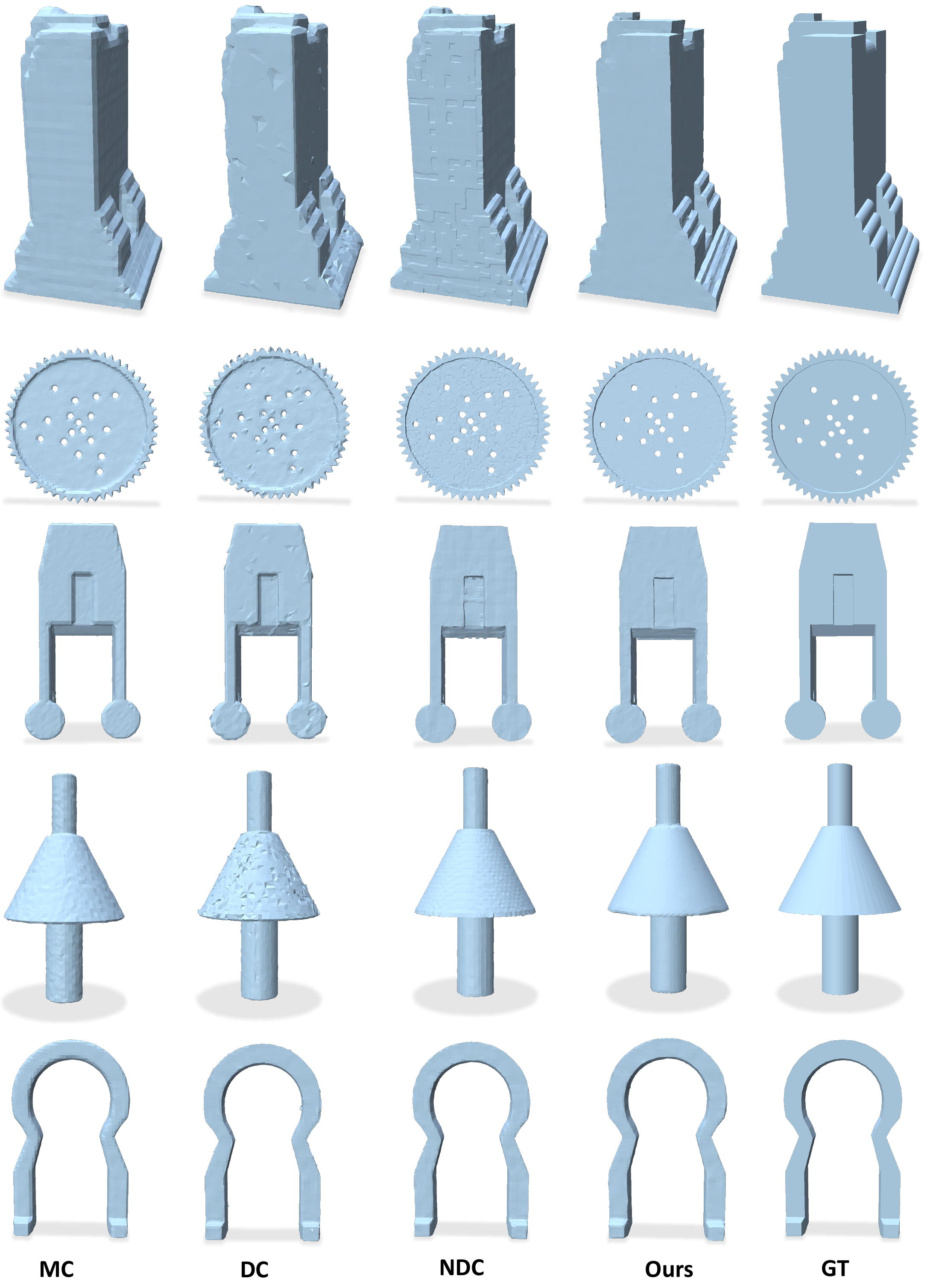}
\caption{Qualitative comparison between reconstructed surfaces from the Thingi10k~\cite{T10k} dataset with input from various learnt methods. First two rows are from SIREN~\cite{sitzmann2020implicit}, second two are from FIN~\cite{tancik2020fourier} and last row is from NGLOD~\cite{takikawa2021nglod}. Our approach produces sharper reconstruction than baselines. }  
\label{fig:LearntSDFQual}
\end{figure*}

\begin{figure*}
\includegraphics[width=1\linewidth]{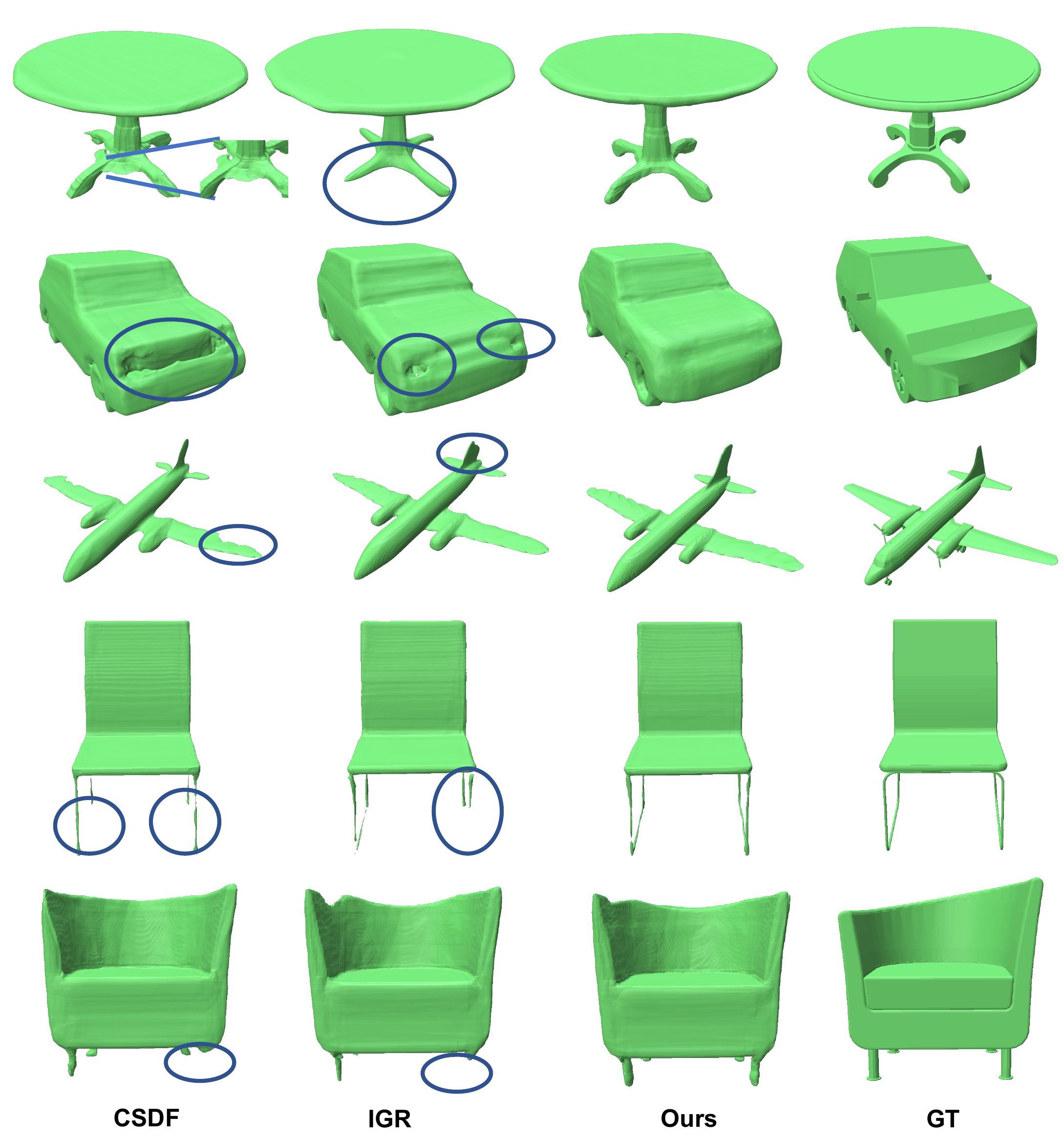}
\caption{Qualitative comparison between reconstructed surfaces from the ShapeNet~\cite{shapenet2015} dataset trained using different implicit surface reconstruction methods. Our approach captures finer details better than the baselines CSDF~\cite{CSDF} and IGR~\cite{IGR}, as indicated by blue circles. }  
\label{fig:sdrQual}
\end{figure*}

In Figure~\ref{fig:LearntSDFQual} we show additional qualitative examples of meshes predicted from the learnt SDF grids of size $128^3$. To show generalization of SDC, we use three Neural Fields, namely SIREN~\cite{sitzmann2020implicit}, Fourier Feature Network~\cite{tancik2020fourier} and NGLOD as our neural field to produce implicit functions. Finally, in Figure~\ref{fig:sdrQual} we show the additional qualitative comparison of our proposed regularization to the relevant baselines.

\pagebreak
{
    \small
    \bibliographystyle{ieeenat_fullname}
    \bibliography{main}
}

\end{document}